\documentclass[10pt,twocolumn,letterpaper]{article}

\usepackage[pagenumbers]{cvpr} 

\def\1n{\mathbf{1}_n}
\def\0{\mathbf{0}}
\def\1{\mathbf{1}}

\definecolor{pink}{rgb}{0.9,0.5,0.5}
\definecolor{purple}{rgb}{0.5, 0.4, 0.8}   
\definecolor{gray}{rgb}{0.3, 0.3, 0.3}
\definecolor{mygreen}{rgb}{0.2, 0.6, 0.2}

\definecolor{greena}{rgb}{0.4, 0.5, 0.1}

\definecolor{bluea}{rgb}{0, 0.4, 0.6}

\definecolor{reda}{rgb}{0.6, 0.2, 0.1}

\newcommand{\cm}[1]{}

\newcommand{\myheading}[1]{\vspace{1ex}\noindent \textbf{#1}}

\newif\ifshowsolution
\showsolutiontrue

\ifshowsolution  

\else

\fi

\newcommand{\Fref}[1]{Fig.~\ref{#1}}
\newcommand{\Tref}[1]{Table~\ref{#1}}

\usepackage{multirow}

\definecolor{cvprblue}{rgb}{0,0,1}
\usepackage[pagebackref,breaklinks,colorlinks,allcolors=cvprblue]{hyperref}
\usepackage{subcaption}
\usepackage{graphicx}
\usepackage{makecell}

\usepackage{amssymb}
\usepackage{pifont}
\newcommand{\cmark}{\ding{51}}%

\usepackage{listings}
\usepackage{xcolor}

\lstdefinestyle{mobind}{
  language=Python,
  basicstyle=\ttfamily\footnotesize,
  commentstyle=\color{blue},
  keywordstyle=\bfseries,
  showstringspaces=false,
  columns=fixed,   
  keepspaces=true,
}

\def\paperID{****} 
\def\confName{CVPR}
\def\confYear{2026}

\title{MoBind: Motion Binding for Fine-Grained IMU–Video Pose Alignment}

\author{
Duc Duy Nguyen \quad Tat-Jun Chin \quad Minh Hoai\\
Australian Institute for Machine Learning, Adelaide University, Adelaide, SA 5000, Australia\\
}

\begin{document}
\maketitle

\begin{abstract}
We aim to learn a joint representation between inertial measurement unit (IMU) signals and 2D pose sequences extracted from video, enabling accurate cross-modal retrieval, temporal synchronization, subject and body-part localization, and action recognition. To this end, we introduce MoBind, a hierarchical contrastive learning framework designed to address three challenges: (1) filtering out irrelevant visual background, (2) modeling structured multi-sensor IMU configurations, and (3) achieving fine-grained, sub-second temporal alignment. To isolate motion-relevant cues, MoBind aligns IMU signals with skeletal motion sequences rather than raw pixels. We further decompose full-body motion into local body-part trajectories, pairing each with its corresponding IMU to enable semantically grounded multi-sensor alignment. To capture detailed temporal correspondence, MoBind employs a hierarchical contrastive strategy that first aligns token-level temporal segments, then fuses local (body-part) alignment with global (body-wide) motion aggregation. Evaluated on mRi, TotalCapture, and EgoHumans, MoBind consistently outperforms strong baselines across all four tasks, demonstrating robust fine-grained temporal alignment while preserving coarse semantic consistency across modalities. Code is available at \url{https://github.com/bbvisual/MoBind}.

\end{abstract}

\section{Introduction}
\label{sec:intro}
Understanding human motion is critical for a wide range of applications, including action recognition~\cite{m_Wang-etal-CVPR20, m_Wang-Hoai-FG18}, sports performance analysis~\cite{mtlaqa, Xu2024FineParserAF, 10.1145/2556288.2557116, yifeng_exrac_aaai_2024, duc_caracount_tpami_2025}, and rehabilitation monitoring~\cite{9207296, Ahad_2019_CVPR_Workshops}. However, human motion is often subtle or hard to sense from a single modality. This limitation can be mitigated by integrating signals from complementary sensing sources, such as video recordings and IMUs. Video can offer rich spatial and semantic information but is sensitive to occlusion, viewpoint changes, and limited frame rates, while IMUs provide precise, temporally dense motion signals but lack visual context, making the captured motion difficult to interpret. To fully leverage the strengths of both modalities, it is essential to develop a joint representation that establishes meaningful correspondence between them—both at the coarse-grained action category level and the fine-grained, sub-second temporal alignment.

\begin{figure}[t]
\centering
\includegraphics[width=0.48\textwidth]{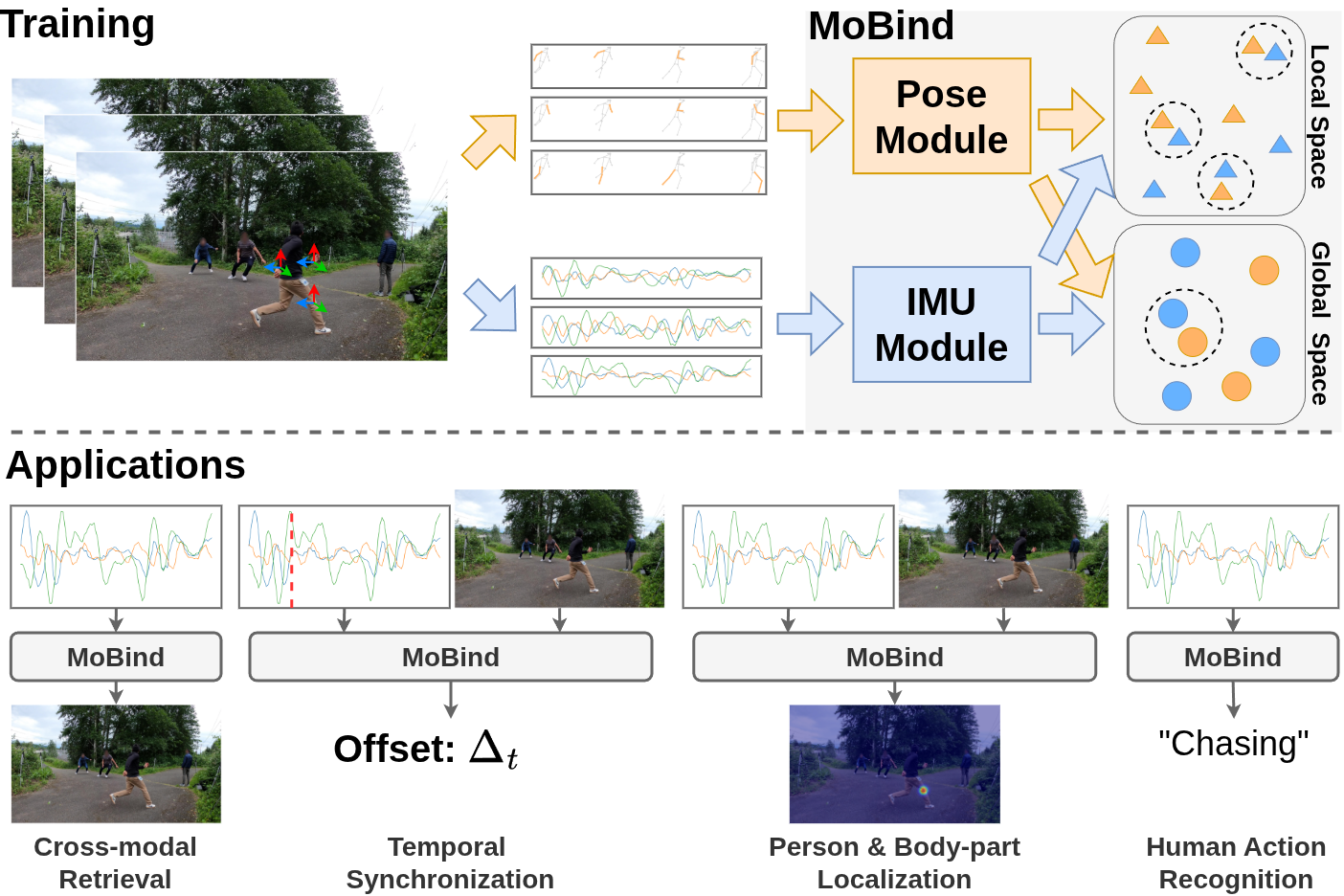}
\vskip -0.1in
\caption{\textbf{Proposed framework for motion binding between IMUs and 2D pose sequence from video.} Contrastive learning is applied at both the local space, aligning each IMU with its corresponding body-part, and the global space, aligning full-body representations. This representation supports several downstream tasks, including cross-modal retrieval, temporal synchronization, subject and body parts localization, and human action recognition.}
\label{fig:teaser}
\vspace{-10px}
\end{figure}

Such a joint representation that aligns IMU signals and video-based body motion would unlock several important capabilities. First, it enables IMU-Video temporal synchronization without cumbersome, explicit calibration procedures. Traditional calibration often requires global timestamps, trigger pulses, or manual alignment—methods that can be technically demanding and error-prone, especially for non-expert users. A learned joint representation enables synchronization based purely on the content of the signals, making it more accessible to collect and use multimodal data. This is important because multimodal data is only truly useful when the different streams are temporally aligned. Second, it supports cross-modal retrieval, where one modality can be used to query relevant information from a database in the other. This is especially valuable in privacy-sensitive scenarios where synchronized video data may be unavailable or restricted, yet similar examples can still be retrieved from the database for visualization and enhanced understanding. Third, it facilitates spatial localization, i.e., associating each IMU with the correct person in the video, which in turn stabilizes tracking in multi-person scenes, and identities can be maintained under occlusions or re-entries. Together, these capabilities make multimodal datasets more reliable to collect, curate, and analyze beyond controlled lab settings.

In practice, much recent work on IMU–vision coupling targets human activity recognition (HAR) by learning a shared embedding space with contrastive objectives~\cite{moon2022imu2clip, girdhar2023imagebind, das2025primus, tan2023egodistill, zhang2025masked}. Typically, each clip or window is projected to a single global embedding, and training relies on matched/mismatched pairs. This design might excel at coarse semantic discrimination (e.g., action categories) but overlooks fine-grained temporal structure: segments that differ only by phase shifts, short lags, or repetition boundaries (i.e., distinct cycles of the same action) collapse to nearby codes. Consequently, the resulting representations become insensitive to true temporal synchrony, a limitation that hinders calibration-free temporal synchronization, sub-second cross-modal retrieval, and spatial localization. This limitation motivates us to develop a joint representation that explicitly models the fine-grained temporal dynamics between IMU and video while keeping the coarse semantics.

In searching for an effective approach to fine-grained IMU–video alignment, we found no prior work that directly targets this setting. We therefore first examined joint audio–video representation learning designed for sub-second alignment~\cite{Chung16a, Afouras20b, 10.5555/3327757.3327874, Fernandez-Labrador_2024_CVPR}, but these techniques do not transfer well to the IMU–video setting. Unlike audio, which often correlates with multiple visual instances in the scene and provides scene-level cues, IMU signals are localized and strictly motion-centric, making most visual background irrelevant. Moreover, IMUs are commonly deployed in multi-sensor configurations, each attached to a different body part, naively concatenating these signals fails to capture their spatial and temporal specificity. Lastly, synchronization cues in audio–video data exist in multiple forms---some are momentary and localized (e.g., door closures, hand claps), while others are more continuous and span broader contexts (e.g., background music aligned with scene changes)~\cite{sparse2022iashin}. Both types are often informative for alignment. In contrast, human motion tends to exhibit continuous and highly repetitive patterns (e.g., walking cycles), producing abundant but highly similar synchronization cues. This repetition can lead to ambiguous alignments, particularly when distinct motion segments appear nearly identical. Taken together, these factors limit the direct applicability of audio–video synchronization techniques and highlight the need for tailored strategies that model sensor-specific dynamics and their alignment with motion-derived visual cues.

In this paper, we introduce a contrastive learning framework illustrated in~\Fref{fig:teaser}, that addresses key limitations of prior works. To focus on motion-relevant cues and reduce irrelevant visual background, we learn a joint representation between IMU signals and video-derived skeletal motion sequences instead of raw pixels. To support multi-sensor IMU setups, we decompose the extracted skeleton into local body-parts and align each with its corresponding IMU signal, enabling structured, semantically grounded association. To achieve fine-grained temporal alignment, we use a hierarchical objective: a local term aligns each IMU–part pair on short segments (sub-second synchrony), and a global term aggregates local features into full-body, multi-IMU embeddings. To retain action-level semantics, we introduce a Masked Token Prediction (MTP) auxiliary task, optimized jointly with the contrastive loss. Evaluated on mRi~\cite{an2022mri}, TotalCapture~\cite{Trumble:BMVC:2017}, and EgoHumans~\cite{Khirodkar_2023_ICCV}, our method consistently outperforms competing approaches across cross-modal retrieval, temporal synchronization, subject/body-part localization, and action recognition.

\section{Related Work}
\label{sec:related}
\subsection{Multi-modal contrastive learning}
In the context of contrastive learning, most IMU work has focused on improving encoders for human activity recognition (HAR)~\cite{haresamudram2021contrastive, 10.1145/3699744, zhang2025masked, 10.1145/3485730.3485937, das2025primus}. However, following the success of CLIP~\cite{pmlr-v139-radford21a} in vision–language modeling, attention has shifted toward learning semantically meaningful cross-modal representations involving IMU signals. IMU2CLIP~\cite{moon2022imu2clip} and ImageBind~\cite{girdhar2023imagebind} align IMU features with the CLIP space, typically via egocentric video, to enable IMU and video retrieval. UniMTS~\cite{zhang2024unimts} maps synthetic IMU from motion data also into the CLIP space to pair with a textual description, enabling zero-shot HAR, while DeSPITE~\cite{despite2025} builds a joint embedding across LiDAR, skeletons, and IMU, optionally incorporating CLIP text when annotations exist, enabling cross-modal retrieval. Although these approaches share the goal of learning joint representations, they generally operate at a global (clip-level), which fails to preserve fine-grained temporal alignment. As our experiments indicate, such formulations lack the precision required for sub-second cross-modal synchronization.

\begin{figure*}[t]
    \centering
    \includegraphics[width=\textwidth]{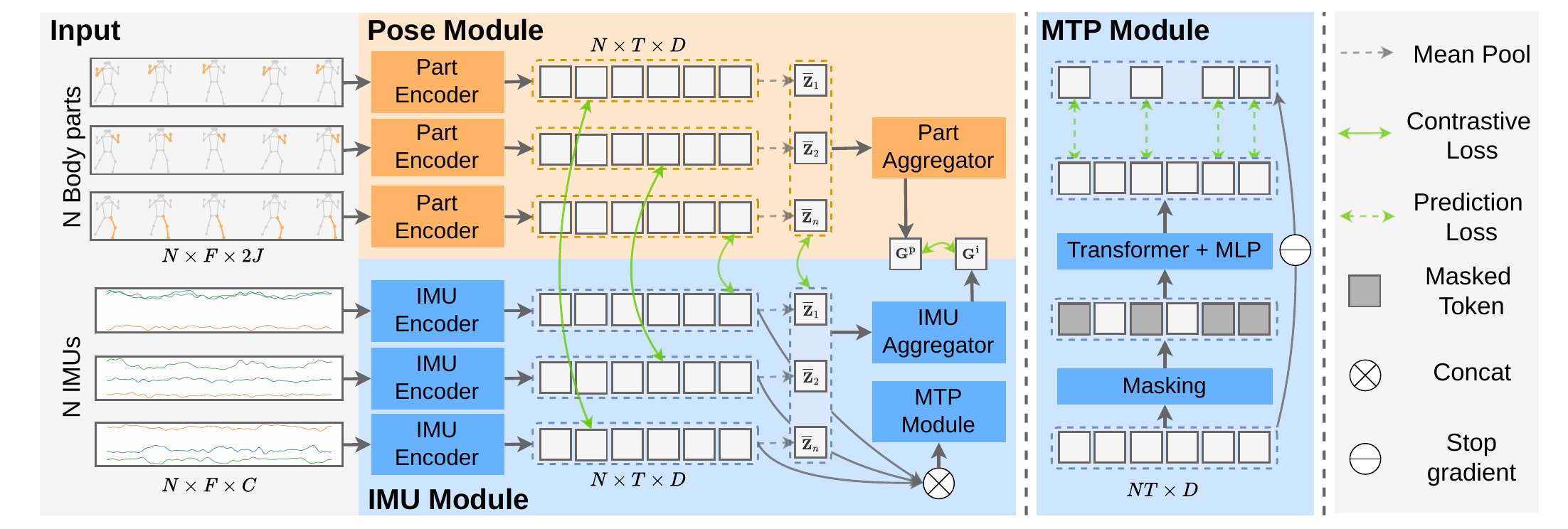}
    \vskip -0.1in
    \caption{\textbf{Overview of the proposed MoBind.} The framework first encodes each IMU stream together with the motion of its corresponding body part, yielding token-level and local-level representations per sensor. These local representations are then aggregated across sensors to form global-level embeddings. The contrastive objective applies at all three levels. In addition, a Masked Token Prediction (MTP) module is used only during training to preserve coarse semantic structure, preventing the model from over-focusing on fine-grained alignment.
} 
    \label{fig:overview}
    \vspace{-10px}
\end{figure*}

\subsection{Temporal synchronization}
Temporal synchronization across modalities has been extensively studied in the audio-visual (AV) domain~\cite{NIPS2000_9f699296, marcheret2015detecting, Chung16a, chung2019perfect, chen2021audio, Fernandez-Labrador_2024_CVPR}. An early approach~\cite{NIPS2000_9f699296} leveraged statistical correlations between MFCCs from audio and mouth appearance in video using correlation to solve AV synchronization. SyncNet~\cite{Chung16a} was among the first to introduce a deep contrastive framework that aligns speech and mouth motion at clip- and frame-level granularity, and DiVAS~\cite{Fernandez-Labrador_2024_CVPR} extended this line with a transformer-based model that is robust to varying frame rates. More recent AV methods~\cite{sparse2022iashin, synchformer2024iashin} replace pure contrastive objectives by tokenizing audio and video and formulating synchronization as temporal offset classification. Inspired by these two paradigms—contrastive alignment and offset classification—we instantiate two strong baselines for IMU-video synchronization by aligning IMU signals with video-extracted skeletal motion. Despite the rich AV literature, IMU-video synchronization remains underexplored. The most closely related work is SyncWISE~\cite{10.1145/3411824}, which estimates temporal offsets by computing correlations between IMU signals and optical-flow derivatives. In contrast, our work provides a fully automatic, learning-based solution that avoids manual gestures or ad-hoc heuristics, making synchronization practical in real-world settings.

\subsection{IMU-to-Subject Association}
The task of associating an IMU with the correct individual in a video has been previously explored. This task is conceptually similar to active speaker detection in audio-visual research~\cite{Chung16a, 9423336, 10.1145/3474085.3475587, Fernandez-Labrador_2024_CVPR}, whose goal is to link audio sources to corresponding visual entities. VIT~\cite{Henschel_2019_CVPR_Workshops} formulates the association as a graph-labeling task, assigning IMU identities to visual tracklets based on orientation consistency. VIPL~\cite{10.1109/IROS45743.2020.9341739} learns a shared visual-inertial feature space using a contrastive loss to match a person's wearable-IMU motion with their video motion. Vi-Fi~\cite{9826015} introduces a recurrent multimodal network that estimates affinity matrices between camera tracks and IMU/device IDs. In contrast to these methods, our approach addresses both spatial association (who is carrying the IMU) and temporal synchronization within a unified learning framework. Moreover, by leveraging the locality of IMU signals, our model also infers body-part attribution-determining not only who is wearing a given IMU, but also where on the body it is worn.


\section{MoBind}
\label{sec:mobi}
This section presents MoBind, an end-to-end framework for learning joint representations between wearable IMU signals and video-based human motion (\Fref{fig:overview}). From the video, skeletal joint sequences are extracted, while raw IMU streams from \(N\) body-mounted sensors are simultaneously processed. Modality-specific modules produce representations trained with a contrastive objective to align synchronized IMU–video pairs and separate mismatched ones. Unlike prior work that compresses inputs into a single global vector, MoBind preserves spatial and temporal structure by capturing body-part motion in \emph{local} representations, aggregating them into a \emph{global} one, and modeling \emph{token-level} temporal dynamics. This hierarchical design can capture fine-grained motion details. However, an exclusive fine-grained focus can under-represent coarse semantics beneficial for HAR. To address this, we add a Masked Token Prediction (MTP) module tailored to IMU inputs, encouraging embeddings to retain high-level semantics while emphasizing temporal granularity. Together, hierarchical alignment and MTP enable the model to learn localized motion patterns and fine-grained dynamics, yielding robust cross-modal alignment and improved HAR performance.

\subsection{Modality-Specific Modules}
\myheading{IMU Module.} The input to the IMU module consists of~$N$ IMU signals, each captured from a sensor mounted on a limb of the human body. Let $\mathbf{X} \in \mathbb{R}^{F \times C}$ denote the data from a specific IMU, where $F$ is the number of frames and~$C$ is the number of sensor channels. We design our IMU encoder $\mathcal{E}_{\text{imu}}$ as a combination of 1D convolutional blocks followed by a Transformer layer, which encodes $\mathbf{X}$ into a sequence of temporal tokens as follows.

A stack of 1D convolutional layers is first applied to $\mathbf{X}$, preserving the temporal resolution $F$ and producing an intermediate feature map $\hat{\mathbf{X}} \in \mathbb{R}^{F \times D}$. This sequence is then divided into $T$ non-overlapping temporal patches $\hat{\mathbf{X}}_1, \ldots, \hat{\mathbf{X}}_T \in \mathbb{R}^{f \times D}$, where $T = F / f$. Each patch is flattened and linearly projected into a $D$-dimensional vector, forming the input for the Transformer layer. The output of the Transformer is a sequence of $T$ temporal tokens  $\mathbf{Z} = [\mathbf{Z}_1, \ldots, \mathbf{Z}_T]$ where $\mathbf{Z}_t \in \mathbb{R}^D$. The {\it local} representation $\overline{\mathbf{Z}} \in \mathbb{R}^D$ for the individual IMU sequence $\mathbf{X}$ is then obtained by applying mean pooling across the temporal dimension of the token sequence.

Let $\{\overline{\mathbf{Z}}_n\}_{n=1}^N$ denote the per-sensor local representations,
$\overline{\mathbf{Z}}_n \in \mathbb{R}^{D}$. We aggregate them by concatenation followed by a
normalization–MLP block:
\[
\mathbf{G}
= \mathrm{MLP}\!\left(\mathrm{LayerNorm}\!\left(\overline{\mathbf{Z}}_{\text{cat}}\right)\right)
\in \mathbb{R}^{D'}.
\]
We refer to $\mathrm{MLP}\circ\mathrm{LayerNorm}$ as the aggregator block, and
$\mathbf{G}$ as the resulting {\it global} representation.

\myheading{Pose Module.} 
The sequence of 2D skeletal joint coordinates is first extracted from an input video. Given the known mounting positions of the IMU sensors, we decompose the full-body motion sequence into $N$ part-specific segments $\mathbf{X}^{\text{part}}_{n} \in \mathbb{R}^{F \times 2J^n}$, each corresponding to a subset of $J^n$ joints. Using the same architectural design as the IMU stream, we employ a body-part encoder $\mathcal{E}_{\text{part}}$, composed of~1D convolutional layers followed by a Transformer. For each body part, the encoder outputs a sequence of $T$ temporal tokens of $D$ dimensions, which encodes the dynamics of the body part's motion. Similar to the IMU stream, these temporal tokens are averaged to produce a local representation for the entire sequence. To derive a global pose representation $\mathbf{G^{\text{part}}}$, we concatenate the \(N\) local vectors and pass them through an aggregator block.

While the number of frames in the video and IMU streams may differ, we ensure that the number of temporal tokens $T$ are the same across the two modalities.

\subsection{Hierarchical Contrastive Alignment} 
To learn fine-grained cross-modal alignment, we adopt contrastive learning in a hierarchical scheme:
(i) \textbf{token-level} alignment, which matches individual temporal tokens across modalities to promote fine-grained correspondence along time (i.e., aligning \(\mathbf{Z}^{\text{imu}}_{t}\) with its counterpart \(\mathbf{Z}^{\text{part}}_{t}\));
(ii) \textbf{local-level} alignment, where each IMU sensor \(n\) is aligned with the motion of its corresponding body part by contrasting \(\overline{\mathbf{Z}}^{\text{imu}}_{n}\) and \(\overline{\mathbf{Z}}^{\text{part}}_{n}\);
and (iii) \textbf{global-level} alignment, where the aggregated IMU representation \({\mathbf{G}}^{\text{imu}}\) is aligned with the global skeletal representation ${\mathbf{G}}^{\text{part}}$.
Before applying the contrastive objective, representations from both modalities are projected into a shared embedding space using modality-specific linear projection layers.

We adopt InfoNCE loss~\cite{Oord2018RepresentationLW} to define global and local contrastive objectives from modality $A$ to modality $B$ for a batch of $K$ paired samples (indexed by $i = 1, \dots, K$) as:
\[
\mathcal{L}_{\text{global}}^{A \to B}
= -\frac{1}{K} \sum_{i=1}^{K}
\log 
\frac{
\exp\left( s\left( {\mathbf{G}}^{A, i}, {\mathbf{G}}^{B, i} \right) / \tau \right)
}{
\sum_{j=1}^{K} \exp\left( s\left( {\mathbf{G}}^{A, i}, {\mathbf{G}}^{B, j} \right) / \tau \right)
},
\]
\[
\mathcal{L}_{\text{local}}^{A \to B}
= -\frac{1}{K} \sum_{i=1}^{K}
\log 
\frac{
\exp\left( s\left( {\overline{\mathbf{Z}}^{A, i}}, {\overline{\mathbf{Z}}^{B, i}} \right) / \tau \right)
}{
\sum_{j=1}^{K} \exp\left( s\left( {\overline{\mathbf{Z}}^{A, i}}, {\overline{\mathbf{Z}}^{B, j}} \right) / \tau \right)
},
\]
where $ s(\cdot, \cdot) $ denotes cosine similarity, and $\tau$ is a learnable temperature parameter. Similarity, the token-level contrastive from modality $A$ to modality $B$ is defined as: 
\[
\mathcal{L}^{A \to B}_{\text{token}} = \frac{-1}{K\,T} \sum_{i=1}^{K} \sum_{t=1}^{T}
\log 
\frac{
\exp\left( s\left( \mathbf{Z}^{A, i}_t, \mathbf{Z}^{B, i}_t \right) / \tau \right)
}{
\sum_{j=1}^{T} \exp\left( s\left( \mathbf{Z}^{A, i}_t, \mathbf{Z}^{B, i}_j \right) / \tau \right)
},
\]
Hence, the contrastive learning objective considering bidirectional alignment between the IMU and Pose modalities, is defined as:
\begin{align*}
\mathcal{L}_{\text{global}} &= \left( \mathcal{L}^{\text{imu} \rightarrow \text{part}}_{\text{global}} + \mathcal{L}^{\text{part} \rightarrow \text{imu}}_{\text{global}} \right)/2, \\
\mathcal{L}_{\text{local}} &= \left( \mathcal{L}^{\text{imu} \rightarrow \text{part}}_{\text{local}} + \mathcal{L}^{\text{part} \rightarrow \text{imu}}_{\text{local}} \right)/2, \\
\mathcal{L}_{\text{token}} &= \left( \mathcal{L}^{\text{imu} \rightarrow \text{part}}_{\text{token}} + \mathcal{L}^{\text{part} \rightarrow \text{imu}}_{\text{token}} \right)/2, \\
\mathcal{L}_{align} &= \lambda_{g} \mathcal{L}_{\text{global}} + \lambda_{l} \mathcal{L}_{\text{local}} + \lambda_{t} \mathcal{L}_{\text{token}},
\end{align*}
where $\lambda_{g}$, $\lambda_{l}$ and $\lambda_{t}$ are weighting coefficients that balance the contributions to the contrastive loss. For brevity, we omit the IMU sensor index $n$ in the local-level and token-level formulations to keep the notation uncluttered.

\setlength{\tabcolsep}{3pt}
\begin{table*}[!t]
\centering
\footnotesize
\begin{tabular}{lcccccccccccccccccc}
\toprule
& \multicolumn{6}{c}{mRi} & \multicolumn{6}{c}{TotalCapture} & \multicolumn{6}{c}{EgoHumans}\\
\cmidrule(lr){2-7} \cmidrule(lr){8-13} \cmidrule(lr){14-19}
Method & \multicolumn{3}{c}{IMU$\rightarrow$Video} & \multicolumn{3}{c}{Video$\rightarrow$IMU} & \multicolumn{3}{c}{IMU$\rightarrow$Video} & \multicolumn{3}{c}{Video$\rightarrow$IMU} & \multicolumn{3}{c}{IMU$\rightarrow$Video} & \multicolumn{3}{c}{Video$\rightarrow$IMU}\\
\cmidrule(lr){2-4} \cmidrule(lr){5-7} \cmidrule(lr){8-10} \cmidrule(lr){11-13} \cmidrule(lr){14-16} \cmidrule(lr){17-19}
& R@1 & R@5 & R@10 & R@1 & R@5 & R@10 & R@1 & R@5 & R@10 & R@1 & R@5 & R@10 & R@1 & R@5 & R@10 & R@1 & R@5 & R@10\\
\midrule
IMU2CLIP & 0.67 & 0.88 & 0.92 & 0.38 & 0.69 & 0.81 & 0.06 & 0.20 & 0.33 & 0.07 & 0.17 & 0.27 & 0.29 & 0.51 & 0.59 & 0.29 & 0.51 & 0.60 \\
DeSPITE & 0.57 & 0.85 & 0.91 & 0.32 & 0.70 & 0.82 & 0.03 & 0.15 & 0.24 & 0.03 & 0.11 & 0.25 & 0.54 & 0.73 & 0.80 & 0.54 & 0.74 & 0.80 \\
SyncNet & 0.77 & 0.94 & 0.97 & 0.75 & 0.92 & 0.95 & 0.51 & 0.87 & 0.94 & 0.54 & 0.86 & 0.93 & 0.74 & 0.80 & 0.93 & 0.71 & 0.89 & 0.93 \\
MoBind & \textbf{0.94} & \textbf{0.99} & \textbf{1.00} & \textbf{0.92 }& \textbf{0.99} & \textbf{0.99} & \textbf{0.87} & \textbf{0.96} & \textbf{0.98} & \textbf{0.68} & \textbf{0.91} & \textbf{0.97} & \textbf{0.83} & \textbf{0.93} & \textbf{0.96} & \textbf{0.83} & \textbf{0.93} & \textbf{0.95} \\
\bottomrule
\end{tabular}
\vskip -0.1in
\caption{
\textbf{Cross-modal retrieval performance on the mRi, TotalCapture and EgoHumans datasets.} We compare our method against prior contrastive learning baselines in both retrieval directions: IMU$\rightarrow$Video and Video$\rightarrow$IMU. Our method consistently outperforms all others across all ranks, demonstrating strong alignment between modalities. This superior performance is particularly critical for downstream tasks that rely on accurate similarity scores between embedded features.
}
\label{tab:retrieval}
\vspace{-10px}
\end{table*}

\subsection{Masked Token Prediction (MTP)}
\label{sec:mtp}
While the hierarchical contrastive objectives capture fine-grained IMU--pose synchrony, they can under-represent the action-level semantics that are useful for downstream action recognition tasks. To complement alignment with a semantics-preserving signal, we introduce a Masked Token Prediction (MTP) auxiliary task that runs in parallel to the alignment branch and is applied to the IMU stream. Let \(\mathbf{Z}^{\text{imu}} \in \mathbb{R}^{N \times T \times D}\) denote the IMU temporal tokens for \(N\) sensors and \(T\) timesteps before local pooling; for brevity, we drop the imu superscript. We sample a mask set \(\mathcal{M} \subset \{1,\ldots,N\}\!\times\!\{1,\ldots,T\}\) with \(|\mathcal{M}|=\lfloor \alpha N T \rfloor\) and replace the selected tokens with a learnable query vector \(\mathbf{q}_{\text{mask}}\), yielding the masked sequence $\mathbf{Z}^{\text{mask}}$. A lightweight Transformer $\mathcal{D}_{\text{mtp}}$ takes the masked sequence $\mathbf{Z}^{\text{mask}}$ as input and uses the unmasked context to predict the missing tokens, $\mathbf{Z}^{\text{pred}}_{n,t} = \big[\mathrm{MLP}\!\left(\mathcal{D}_{\text{mtp}}(\mathbf{Z}^\text{mask})\right)\big]_{n,t} \quad \text{for } (n,t)\in\mathcal{M}$. MTP loss is the mean-squared error over masked positions:
\begin{equation}
\mathcal{L}_{\text{mtp}}
=\frac{1}{|\mathcal{M}|}\sum_{(n,t)\in\mathcal{M}}\left\|\mathbf{Z}^{\text{pred}}_{n,t}-\mathbf{Z}_{n,t}\right\|_2^2
\end{equation}
We jointly optimize MTP with the alignment objective using a scalar weight \(\lambda_{\text{mtp}}\):
$
\mathcal{L} \;=\; \mathcal{L}_{\text{align}} \;+\; \lambda_{\text{mtp}}\,\mathcal{L}_{\text{mtp}}$.

\section{Experiments}
\label{sec:exps}
MoBind provides a joint representation that supports a range of downstream tasks: cross-modal retrieval, temporal synchronization, subject and body-part localization, and action recognition. We first introduce the evaluation datasets, then—for each downstream task—detail the setup, metrics, baselines, and experiment results. Finally, we present ablation studies that quantify the contribution of key design choices and evaluate robustness to real-world conditions via simulated sensor dropouts.

\subsection{Datasets \& Training}
We perform experiments on three multimodal datasets: mRi~\cite{an2022mri}, TotalCapture~\cite{Trumble:BMVC:2017}, and EgoHuman~\cite{Khirodkar_2023_ICCV}, where EgoHumans contains multi-person scenes. MoBind is trained with fixed $5\,\mathrm{s}$ windows, and the same window length is used for retrieval, localization, and action recognition. 
To reflect realistic deployment conditions, we use estimated 2D keypoints throughout all experiments: pose sequences are predicted with MMPose software~\cite{mmpose2020} using RTMPose~\cite{https://doi.org/10.48550/arxiv.2303.07399}. We adopt subject-wise splits for mRi and TotalCapture, and a non-overlapping scene split for EgoHumans.

Using a $5\,\mathrm{s}$ input window, each segment yields $T=25$ temporal tokens. We use 256-dimensional embeddings for both local $(D)$ and global $(D')$ representations. Loss weights are $\lambda_{g}=1.0$, $\lambda_{l}=1.0$, $\lambda_{t}=0.5$, and $\lambda_{\text{mtp}}=0.3$. The MTP auxiliary task uses a masking ratio $\alpha=0.75$. MoBind is trained with Adam optimizer~\cite{Kingma2015AdamAM} (learning rate $1\times10^{-4}$, batch size $1356$) using early stopping based on the R@1 retrieval score on the validation set, with a patience of 500 epochs. Training runs on a single NVIDIA GeForce RTX 5090 and takes about 2.5 hours per run. More details can be found in the supplementary material.

\subsection{Cross-Modal Retrieval}

\myheading{Experiment setup.} Given a short segment from one modality, the goal is to retrieve the corresponding moment in the other. For each batch of IMU–video data, we first extract 2D pose sequences from the videos, then embed all IMU and pose segments, and compute cosine similarity scores for every query–reference pair. These scores form a similarity matrix, where the entry at row $i$ and column $j$ represents the similarity between query IMU $i$ and reference pose sequence $j$. For IMU-to-Video retrieval, we sort each query's scores in descending order to rank pose references, with higher scores indicating better matches. The process is analogous for Video-to-IMU retrieval, where pose segments act as queries and IMU segments as references.

\myheading{Baselines.} We compare MoBind against strong contrastive IMU baselines: IMU2CLIP~\cite{moon2022imu2clip}, DeSPITE~\cite{despite2025}, and SyncNet~\cite{Chung16a} (originally for audio–visual synchronization, here adapted to the IMU–pose setting). 
We adopt a multi-positive contrastive mechanism, where a single pose sequence can correspond to multiple IMU sensors, and the final IMU embedding is obtained by averaging their representations. All experiment results are reported under this setting. The concatenated variant and additional implementation details for MoBind and the baselines are provided in the supplementary material. We evaluate bidirectional retrieval (IMU$\rightarrow$Video and Video$\rightarrow$IMU) using the Recall@k metrics, with $k \in \{1, 5, 10\}$.

\myheading{Results.} \Tref{tab:retrieval} compares retrieval on mRi, TotalCapture, and EgoHumans. MoBind consistently outperforms all baselines in both retrieval directions, with especially large gains on mRi and TotalCapture. On mRi, among the R@1 errors, the share whose top-1 impostor belongs to the same action class is 0.79 (IMU2CLIP), 0.76 (DeSPITE), and 0.75 (SyncNet). This confusion indicates that these methods compress sequences into a single global vector that primarily encodes coarse semantics, useful for action recognition but insufficient for instance-level alignment. On TotalCapture, the effect is starker: the median hard-negative margin (cosine similarity of the true match minus that of the hardest impostor) is -0.14, -0.11, and -0.14 for IMU2CLIP, DeSPITE, and SyncNet, respectively, whereas MoBind attains +0.10. Thus, for the baselines, the hardest impostor is typically more similar than the true match, while MoBind maintains a positive gap. These findings are consistent with the quantitative gains and indicate that global-only embeddings struggle in TotalCapture's highly dynamic scenes, where whole-body motion overwhelms limb-level cues, hindering alignment to the worn IMUs. By first aligning at the local body-part level and then aggregating globally, MoBind preserves local information and remains robust. Qualitative IMU-Video retrieval examples are shown in \Fref{fig:qual_retrieval}.

\newcommand{\figheight}{1.016cm}
\begin{figure}[t]
\centering
\begin{minipage}[t]{.49\linewidth}
  \includegraphics[width=\linewidth]{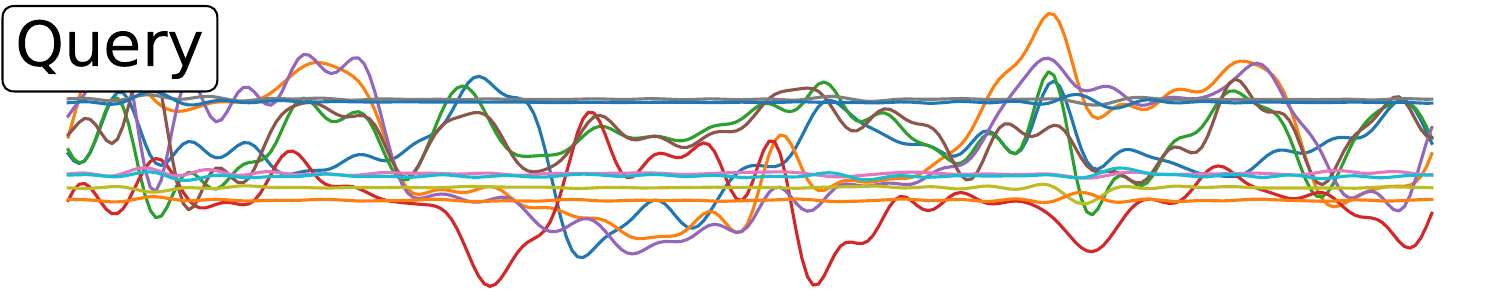}\par
  \includegraphics[width=\linewidth]{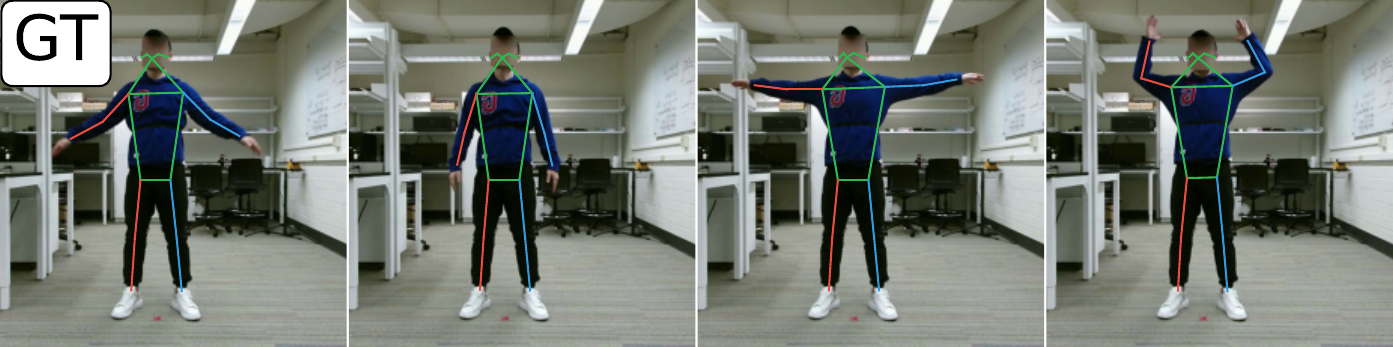}\par
  \includegraphics[width=\linewidth]{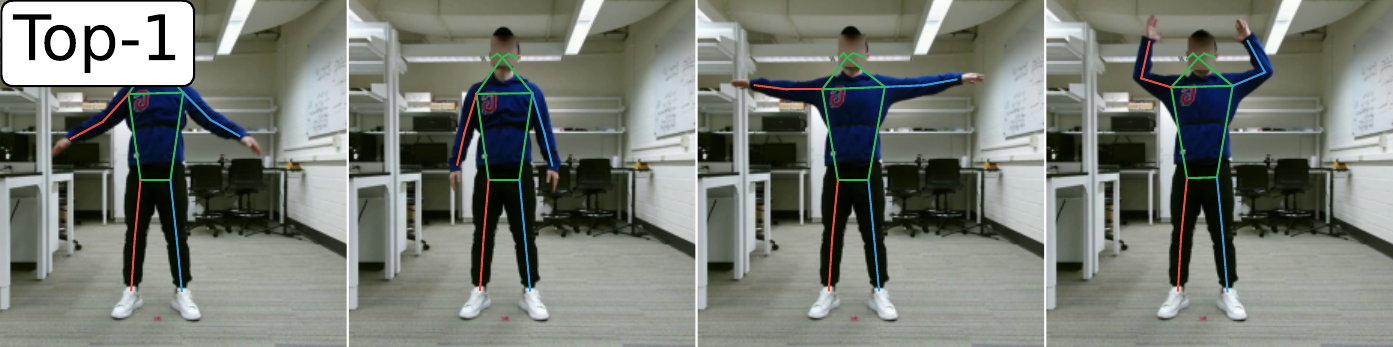}\par
  \includegraphics[width=\linewidth]{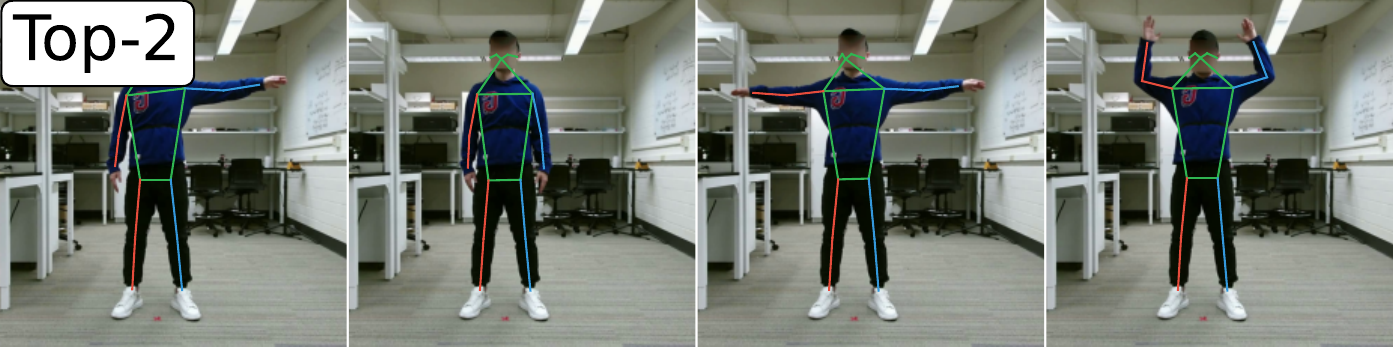}\par
  \includegraphics[width=\linewidth]{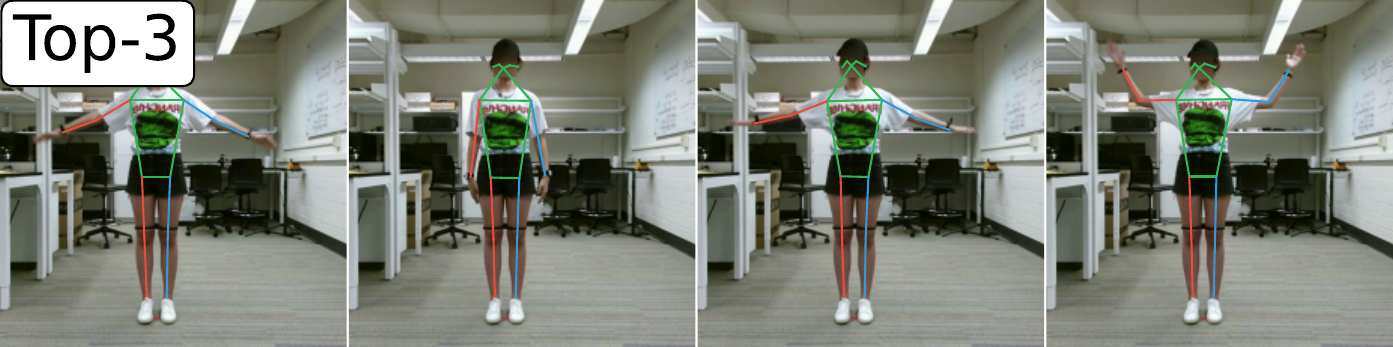}
\end{minipage}
\hfill
\begin{minipage}[t]{.49\linewidth}
  \includegraphics[width=\linewidth]{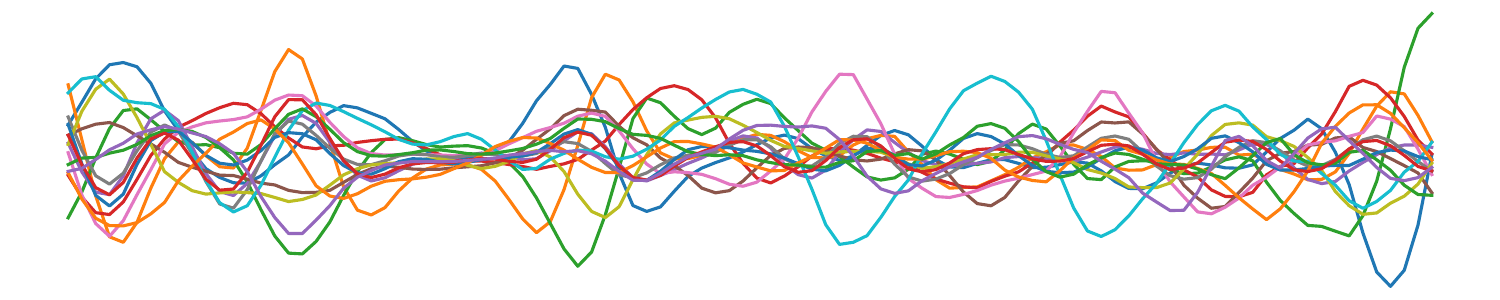}\par
  \includegraphics[height=\figheight, width=\linewidth]{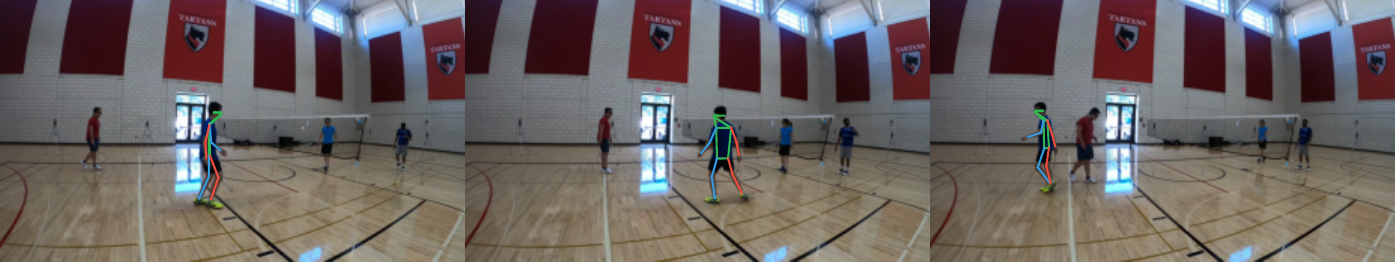}\par
  \includegraphics[height=\figheight, width=\linewidth]{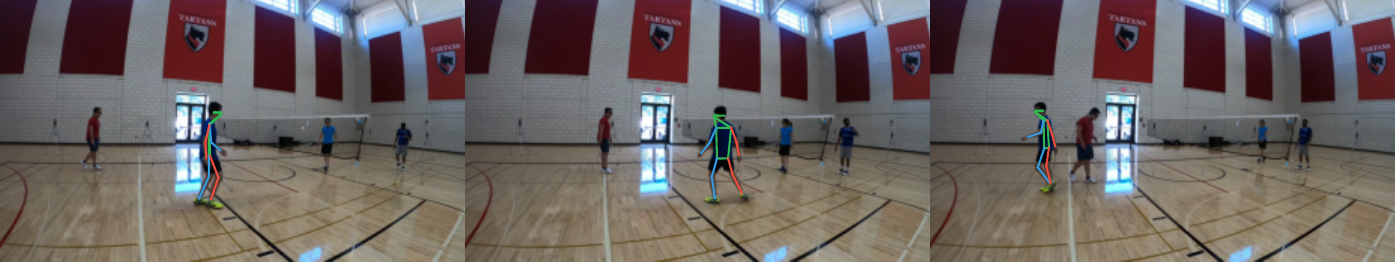}\par
  \includegraphics[height=\figheight,width=\linewidth]{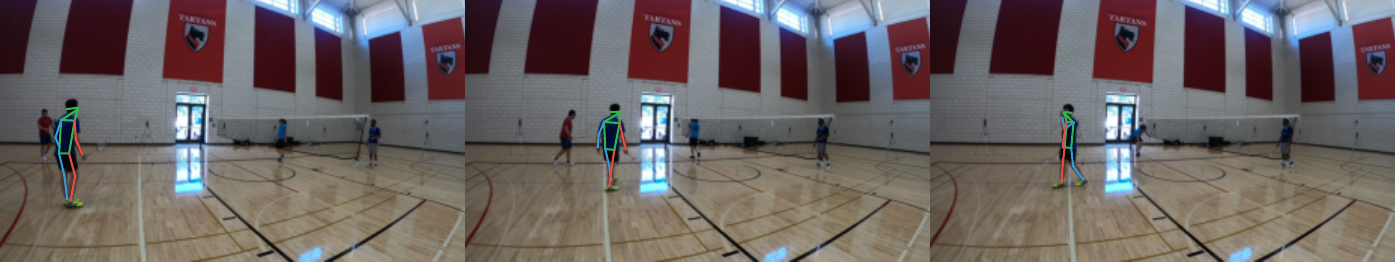}\par
  \includegraphics[height=\figheight,width=\linewidth]{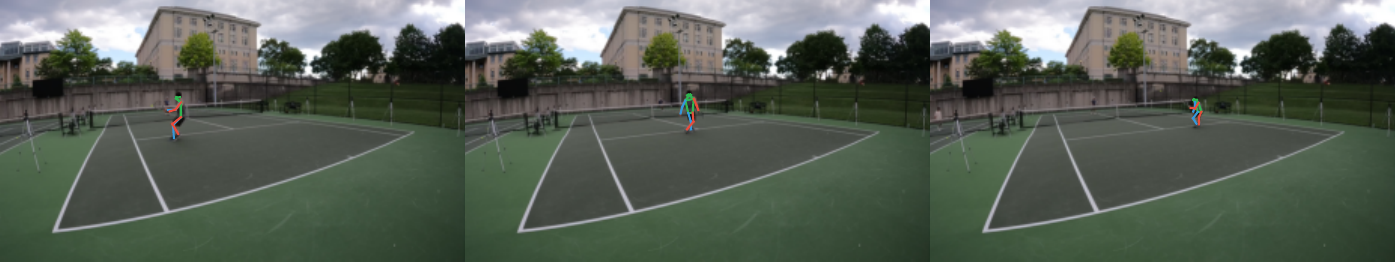}
\end{minipage}
\vskip -0.1in
\caption{\textbf{IMU\(\rightarrow\)Video retrieval results} on mRi (left) and EgoHumans (right). Each example shows the query IMU signal, its corresponding ground-truth video segment, and the top three retrieved video segments. Our method successfully retrieves the ground-truth segment, and the other top-ranked results are also visually similar to the ground truth, demonstrating robust cross-modal alignment.}
\label{fig:qual_retrieval}
\vspace{-10px}
\end{figure}

\subsection{Temporal Synchronization}
\myheading{Experiment Setup.}  Given paired IMU-video sequences, the goal of this task is to estimate the temporal offset between the two modalities. For this experiment, we construct a test set of IMU-video sequences of $20\mathrm{s}$ in duration and simulate misalignment by introducing random lags uniformly sampled from $[-7, 7]\mathrm{s}$ between the modalities. We first extract the pose sequence from the video and then divide each modality into \(N\) overlapping temporal windows, each with the length corresponding to duration of the temporal windows used for training MoBind ($5\mathrm{s}$), resulting in paired segments \((\mathbf{X}^{\text{imu}}_n, \mathbf{X}_n^{\text{part}})\). Each segment is passed through MoBind to obtain modality-specific feature embeddings $(\mathbf{G}^{\text{imu}}_n, \mathbf{G}_n^{\text{part}})$. Next, we compute a pairwise similarity matrix \(D \in \mathbb{R}^{N \times N}\) where each entry \(D_{p,q}\) is the cosine similarity between the \(p\)-th IMU window and the \(q\)-th pose window. For each IMU window \(p\), we retrieve the top-\(k\) most similar pose windows \(q\) based on \(D_{p,q}\), and vice versa. Each matched pair \((p, q)\) yields an offset vote: $\Delta_{p \rightarrow q} = q - p$ with the associated similarity score \(D_{p,q}\) used as the vote's weight. All offset votes from both retrieval directions are aggregated into a weighted histogram, where each offset bin accumulates the similarity scores of matching pairs that fall into that bin. The final estimated temporal offset $\hat{\delta}$ is obtained as the bin with the highest total weight:
\[
\hat{\delta} = \operatorname*{arg\,max}_{\Delta} \sum_{(p,q)\,:\,q-p=\Delta} D_{p,q}
\]

\myheading{Baselines.} In addition to the contrastive baselines, and inspired by recent learning-based audio–video synchronization methods~\cite{sparse2022iashin, synchformer2024iashin}, we design an offset-classification baseline, which we call IMUSync. We also include SyncWISE~\cite{10.1145/3411824}, a correlation-based method. Performance is reported using mean absolute error (MAE) in seconds and accuracy within a 200ms tolerance.

\setlength{\tabcolsep}{3pt}
\begin{table}[]
\small
\centering
\begin{tabular}{lcccccc}
\toprule
\multirow{2}{*}[-1ex]{Method} & \multicolumn{2}{c}{mRi} & \multicolumn{2}{c}{TotalCapture} & \multicolumn{2}{c}{EgoHumans}\\ 
\cmidrule(lr){2-3} \cmidrule(lr){4-5} \cmidrule(lr){6-7}
& \multicolumn{1}{c}{MAE~$\downarrow$} & \multicolumn{1}{c}{Acc~$\uparrow$} & \multicolumn{1}{c}{MAE~$\downarrow$} & \multicolumn{1}{c}{Acc~$\uparrow$} & \multicolumn{1}{c}{MAE~$\downarrow$} & \multicolumn{1}{c}{Acc~$\uparrow$}\\
\midrule
Random guess & 10.72 & 0.00 & 10.06 & 0.00 & 10.18 & 0.01 \\
SyncWISE & 3.31  & 0.04 & 4.07 & 0.02 & 3.68 & 0.02\\
IMU2CLIP & 1.17 & 0.70 & 2.32 & 0.13 & 3.13 & 0.44\\
DeSPITE & 1.42 & 0.63 & 2.38 & 0.16 & 2.58 & 0.58\\
SyncNet & 0.89 & 0.76 & 1.85 & 0.21 & 2.93 & 0.39\\
IMUSync & 0.72 & 0.75 & 0.96 & 0.71 & 1.01 & 0.82 \\
MoBind & \textbf{0.47} & \textbf{0.88} & \textbf{0.05} & \textbf{0.98} & \textbf{0.04} & \textbf{1.00}\\
\bottomrule 
\end{tabular}
\vskip -0.1in
\caption{
\textbf{Synchronization results on three datasets.} All models are evaluated on 20-second videos with random temporal offsets sampled uniformly from $[-7, 7]$ seconds. Top-$k$ retrieval is performed with $k = 5$. Our method significantly outperforms baselines in both mean absolute error (MAE, in seconds) and accuracy (Acc, within a 200ms tolerance) across all datasets.
}
\label{tab:res_sync}
\end{table}

\begin{figure}[t]
\centering
\includegraphics[width=0.9\linewidth]{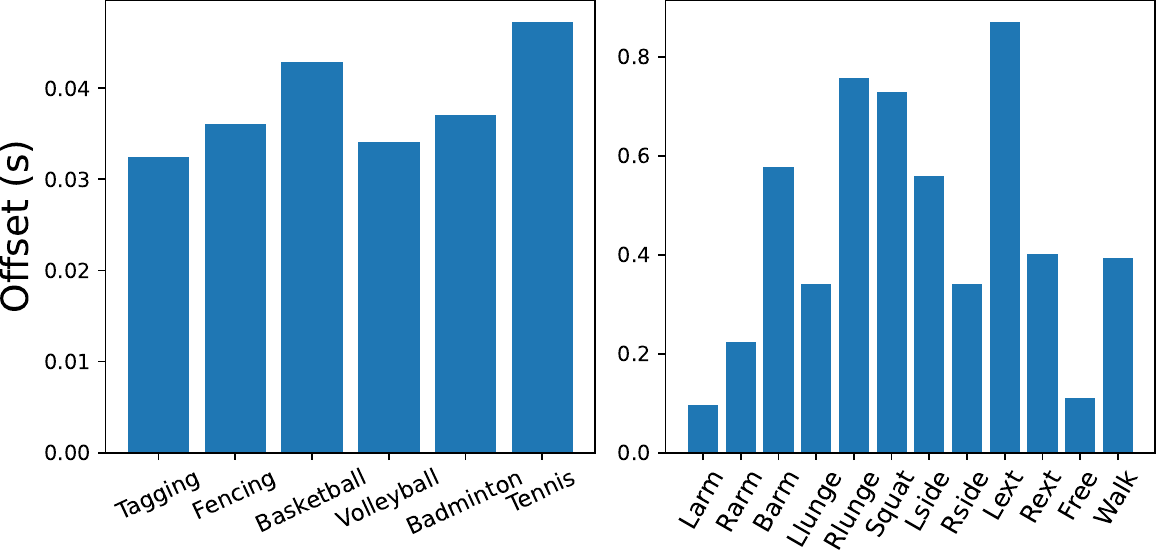}
\vskip -0.1in
\caption{
\textbf{Per-action synchronization accuracy} on EgoHumans (left) and mRi (right). MoBind achieves sub-50ms error on all EgoHumans actions and under 1s on all mRi actions, despite the challenges posed by repetitive movements and near-duplicate segments. Results confirm MoBind's robustness across diverse motion types and environments.
}
\label{fig:sync}
\vspace{-10px}
\end{figure}

\myheading{Results.}
\Tref{tab:res_sync} demonstrates MoBind’s ability to correct temporal misalignment between video and IMU streams. We perform bidirectional retrieval with $k{=}5$ using $20\,\mathrm{s}$ clips, partitioned into overlapping $5\,\mathrm{s}$ segments with a $0.2\,\mathrm{s}$ stride (yielding 76 segments per clip). This setup captures fine-grained temporal correspondences, moving beyond coarse retrieval to enable sub-second alignment.


MoBind consistently achieves accurate synchronization across diverse real-world scenarios, including rehabilitation settings (mRi), dynamic actor-driven motions (TotalCapture), and complex multi-person activities (EgoHumans). Per-action synchronization results are visualized in \Fref{fig:sync}, showing MoBind's robust offset estimation across a wide range of action categories. On EgoHumans, the synchronization error remains below 50ms for all actions. On mRi, while performance is slightly lower due to the presence of repetitive exercises that produce many near-duplicate segments and hard negatives, MoBind still maintains errors under one second for all categories.



Importantly, MoBind supports synchronization of IMU and video sequences of arbitrary lengths. Longer input durations yield higher accuracy: on the challenging mRi dataset where many action classes involve repetitive movements, synchronization accuracy reaches 97\% with 2-minute clips and achieves perfect alignment (100\%) with 3-minute clips.

\def\subFigSz{0.33\linewidth} 
\begin{figure}[t] 
\centering
    \includegraphics[width=\subFigSz]{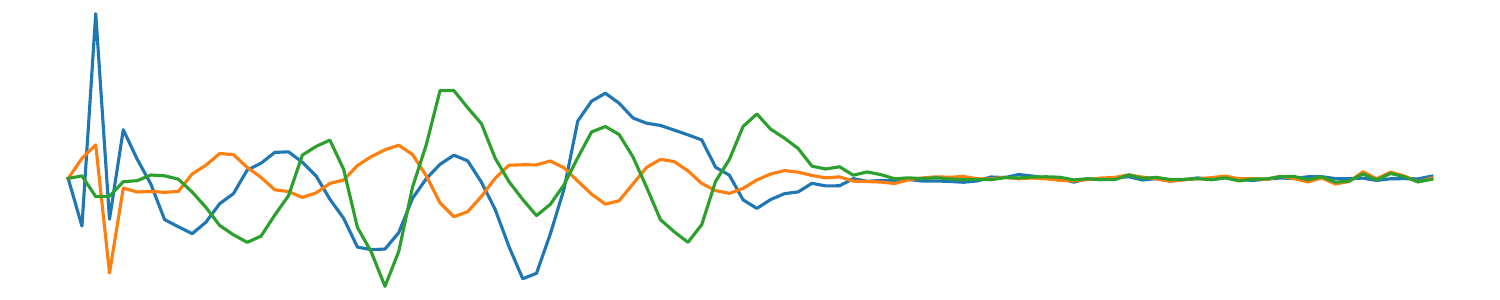}\hfill
    \includegraphics[width=\subFigSz]{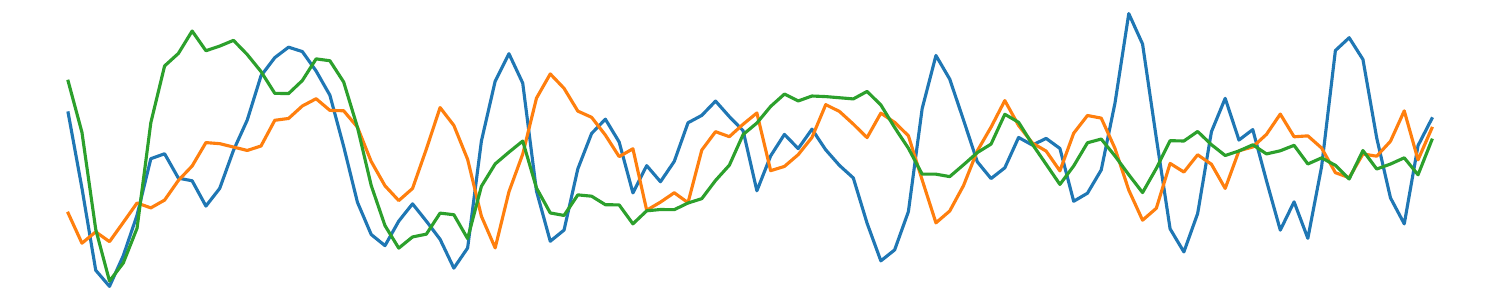}\hfill
    \includegraphics[width=\subFigSz]{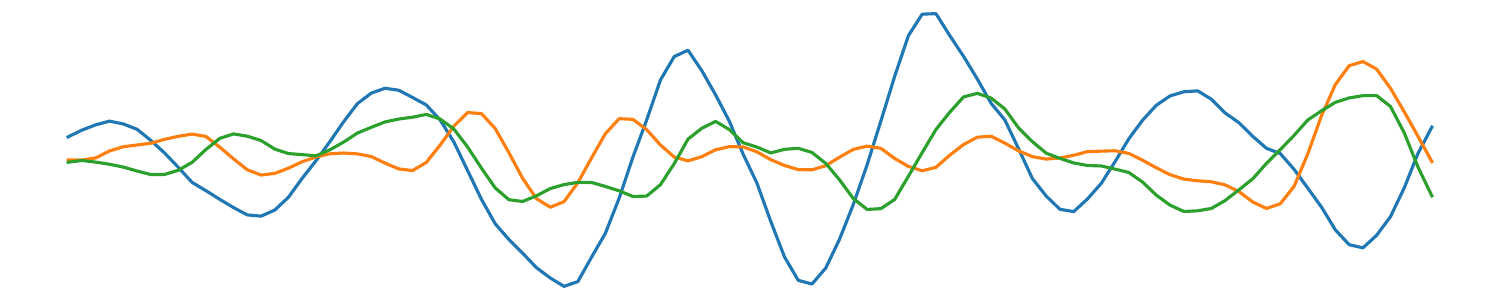}
    \includegraphics[width=\subFigSz]{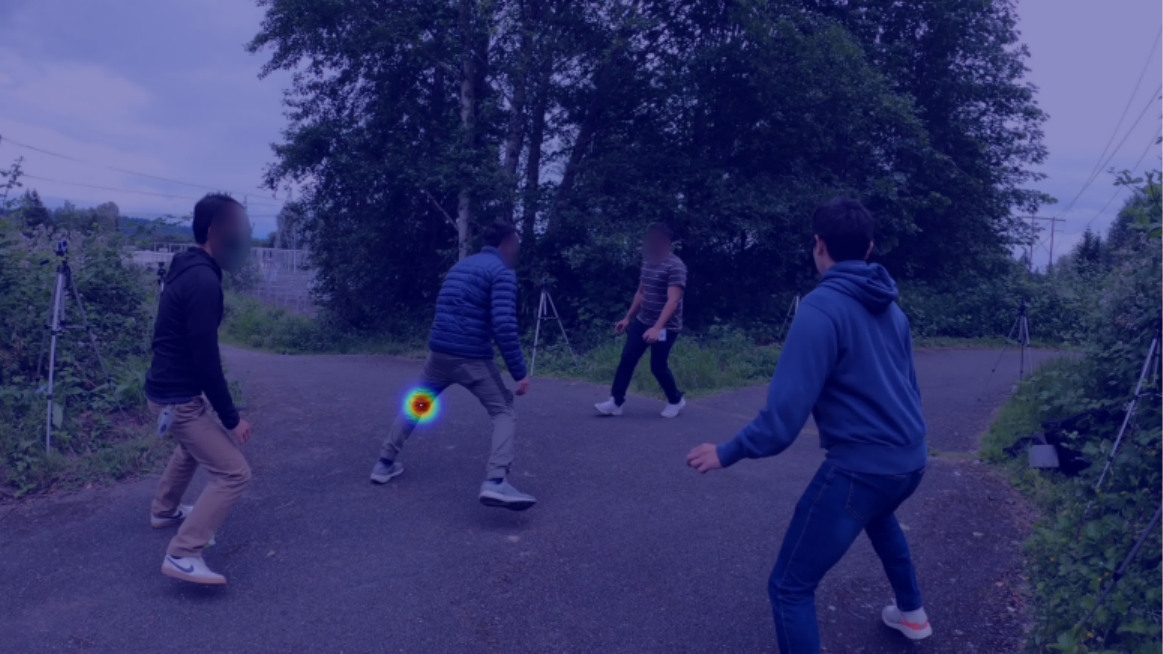}\hfill
    \includegraphics[width=\subFigSz]{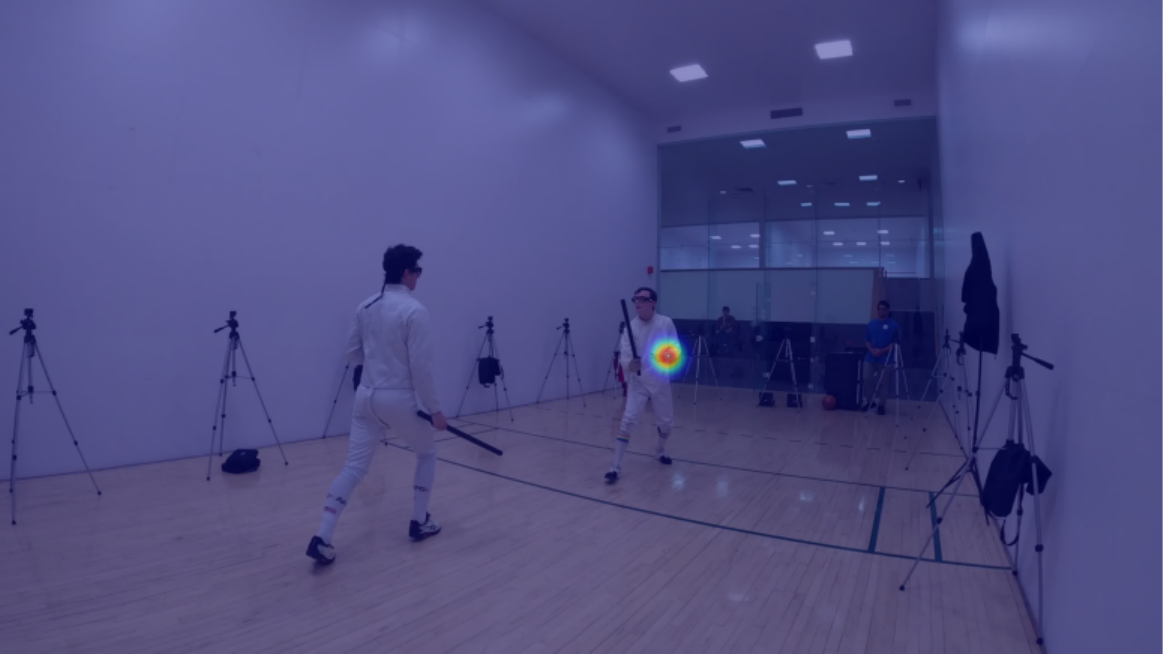}\hfill
    \includegraphics[width=\subFigSz]{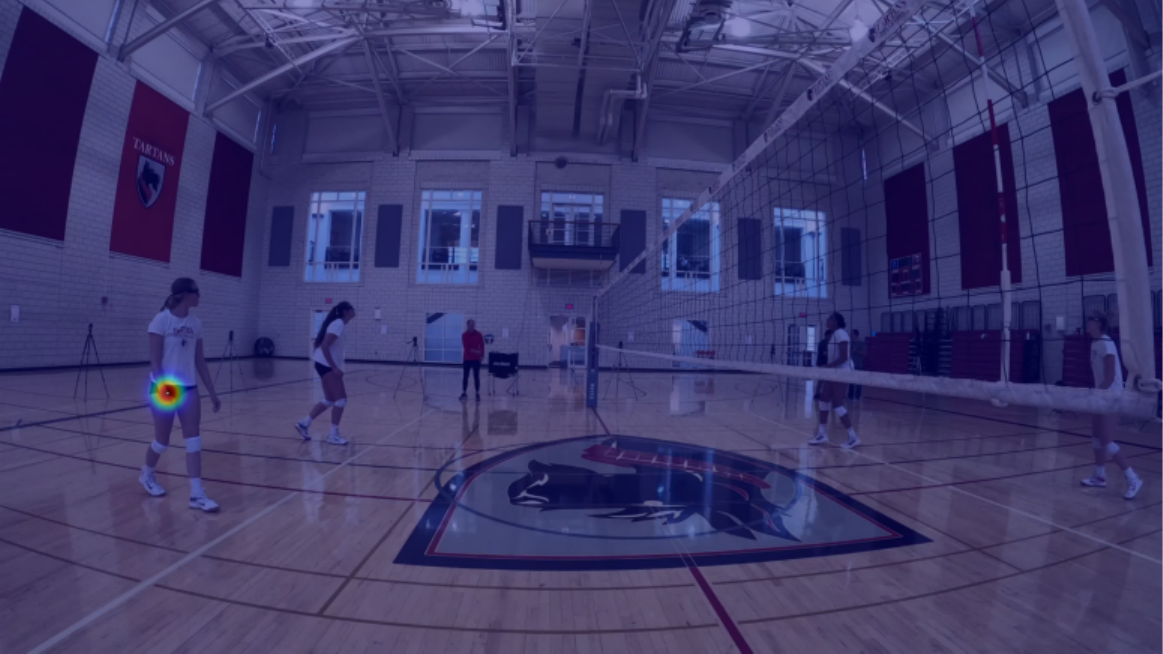}\\
\vskip -0.1in
\caption{
\textbf{Examples of body-part localization} on EgoHumans. Each column shows the query IMU (top) and the predicted body part with the highest similarity score (bottom), demonstrating accurate identification of sensor placement.}
\label{fig:spatial}
\vspace{-5px}
\end{figure}

\setlength{\tabcolsep}{5pt}
\begin{table}[t]\footnotesize
\centering
\begin{tabular}{lccccc}
\toprule
Method & IMU2CLIP & DeSPITE & SyncNet & VIPL & MoBind \\ 
\midrule
Acc & 0.9446 & 0.9759 & 0.9749 & 0.9014 & \textbf{0.9812} \\
F1-score & 0.9430 & 0.9743 & 0.9748 & 0.8933 & \textbf{0.9801}\\
\bottomrule 
\end{tabular}
\vskip -0.1in
\caption{{\bf IMU-to-person identification in multi-person scenes} from EgoHumans. This experiment evaluates who wears the IMU sensor, and MoBind achieves the highest accuracy and F1 score.}
\label{tab:spatial}
\vspace{-10px}
\end{table}

\subsection{Subject and Body-part Localization}
\myheading{Experiment setup.} Given a synchronized IMU segment and a multi-person video, we identify the IMU wearer by comparing the IMU's global embedding to each person's global pose embedding generated by MoBind, using cosine similarity. The person with the highest similarity is then assigned as the IMU wearer. For body-part association, we apply the same approach using local (body-part) embeddings: we compare the IMU's local embedding with all candidate body-part embeddings across individuals and assign the pair with the highest similarity.

\myheading{Results.} \Tref{tab:spatial} shows MoBind's person localization performance on EgoHumans, achieving the highest accuracy and F1-score compared to baselines, including VIPL~\cite{10.1109/IROS45743.2020.9341739}.

\begin{figure}[t] 
\centering
\includegraphics[width=\linewidth]{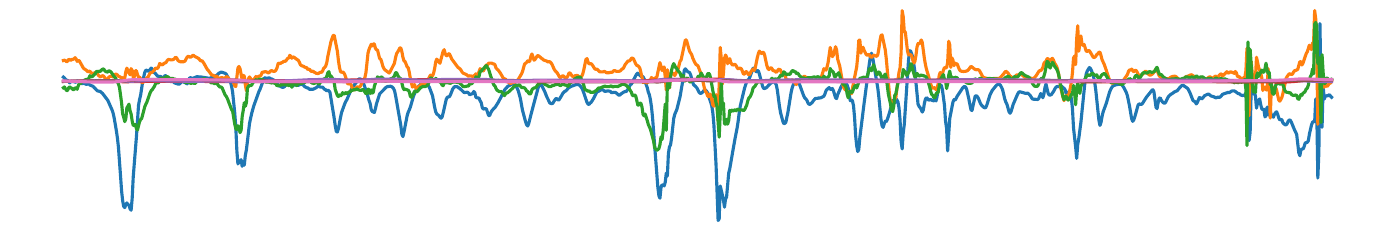}\par
\includegraphics[width=\linewidth]{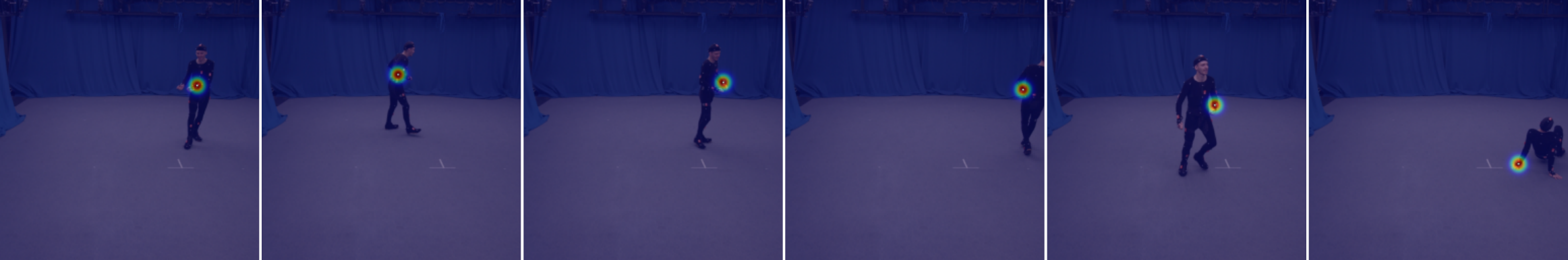}\par
\includegraphics[width=\linewidth]{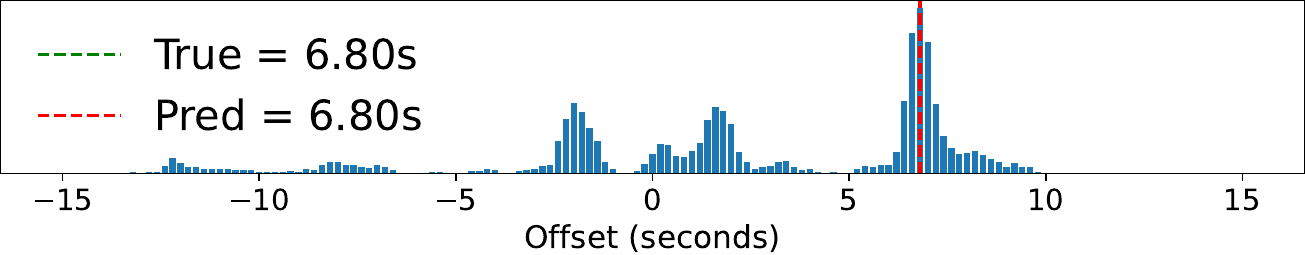}
\vskip -0.1in
\caption{
\textbf{Results for the challenging combined task of temporal synchronization and spatial localization.} The query $20\mathrm{s}$ IMU signal from the left wrist precedes the video by $6.8\,\text{s}$. MoBind accurately recovers this offset using a weighted histogram and simultaneously localizes the signal to the correct body part, demonstrating robustness to dynamic and highly repetitive motion.
\label{fig:spatial_sync}}
\vspace{-5px}
\end{figure}

MoBind incorporates local contrastive learning in addition to global alignment, enabling it to identify not only the person wearing an IMU sensor but also the specific body part to which the sensor is attached. As shown in~\Fref{fig:spatial}, given a query signal from a specific IMU, our method correctly assigns it to the corresponding body part of the subject in the video. Across datasets, MoBind achieves body-part localization accuracies of 0.81 (mRi), 0.57 (TotalCapture), and 0.63 (EgoHumans). Similair to synchronize, we can also improve this accuracy with longer input window. Moreover, this capability naturally extends to the combined task of joint spatial grounding and temporal synchronization, as illustrated in~\Fref{fig:spatial_sync}. MoBind not only detects temporal misalignment but also correctly grounds the query IMU to the left wrist, demonstrating the framework’s flexibility and practical utility.

\setlength{\tabcolsep}{5pt}
\begin{table}[]
\small
\centering
\begin{tabular}{lcccc}
\toprule
\multirow{2}{*}[-1ex]{Method} & \multicolumn{2}{c}{mRi} & \multicolumn{2}{c}{TotalCapture}\\ 
\cmidrule(lr){2-3} \cmidrule(lr){4-5}
& \multicolumn{1}{c}{Finetune} & \multicolumn{1}{c}{1-NN} & \multicolumn{1}{c}{Finetune} & \multicolumn{1}{c}{1-NN} \\
\midrule
UniMTS & 0.95 & 0.36 &  0.66 & 0.53\\
ImageBIND & 0.95 & 0.40 & \textbf{0.72} & 0.47\\
EVI-MAE & 0.54 & 0.43 & 0.63 & 0.66\\
Primus & 0.82 & 0.80 & 0.57 & 0.67\\
IMU2CLIP & 0.79 & 0.85 & 0.60 & 0.68 \\
DeSPITE & 0.79 & \textbf{0.87} & 0.50 & 0.53\\
SyncNet & 0.97 & 0.85 & 0.68 & 0.63\\
MoBind & \textbf{0.98} & \textbf{0.87} & \textbf{0.72} & \textbf{0.71}\\
\bottomrule 
\end{tabular}
\vskip -0.1in
\caption{
 \textbf{Human activity recognition results under finetuning and 1-NN setting on the mRi and TotalCapture.} For consistency, fine-tuning results are reported at the 5\textsuperscript{th} epoch.
}
\label{tab:har}
\vspace{-10px}
\end{table}

\subsection{Human Action Recognition}
\myheading{Experiment Setup.}
Beyond fine-grained temporal alignment, we evaluate the semantic quality of MoBind by applying it to human action recognition. We consider two classifiers:
(1) 1-Nearest Neighbor (1-NN)~\cite{10.1145/3474085.3475307, Guo_Liu_Chen_Liu_Wang_Ding_2022, 10.1609/aaai.v37i1.25127}, where each action is predicted based on the cosine similarity to the nearest labeled example in the representation space;
(2) Fine-tuned classifier, where a linear classification head is trained jointly with the IMU encoder.


We benchmark MoBind against the baseline representations used throughout this work, as well as several recent state-of-the-art IMU encoders: UniMTS~\cite{zhang2024unimts}, ImageBind~\cite{girdhar2023imagebind}, EVI-MAE~\cite{zhang2025masked}, and Primus~\cite{das2025primus}. These additional encoders are designed exclusively for IMU signals and are suitable for human activity recognition, but not for cross-modal synchronization or retrieval. Therefore, they were not included in the previous experiments.


\setlength{\tabcolsep}{3pt}
\begin{table}[]
\centering
\begin{tabular}{ccccccccc}
\toprule
\multicolumn{3}{c}{Combinations} & \multicolumn{2}{c}{Retrieval} & \multicolumn{2}{c}{Sync.} & \multirow{2}{*}[-1ex]{\makecell{Locali\\zation}}\\ 
\cmidrule(lr){1-3} \cmidrule(lr){4-5} \cmidrule(lr){6-7}
\multicolumn{1}{c}{global} & \multicolumn{1}{c}{local} & \multicolumn{1}{c}{token} & \multicolumn{1}{c}{I $\rightarrow$ V} & \multicolumn{1}{c}{V $\rightarrow$ I} & \multicolumn{1}{c}{MAE} & \multicolumn{1}{c}{Acc}\\
\midrule
\cmark &   & & 0.34 & 0.31 & 0.74 & 0.81 & 0.22\\
\cmark & \cmark &  & 0.77 & 0.78 & 0.49 & 0.86 & 0.75\\
\cmark & \cmark & \cmark & \textbf{0.94} & \textbf{0.92} & \textbf{0.47} & \textbf{0.88} & \textbf{0.81}\\
\bottomrule 
\end{tabular}
\vskip -0.1in
\caption{
 \textbf{Ablation studies on contrastive objectives on mRi.} Results show consistent gains across all tasks as each contrastive level is added, highlighting the effectiveness of the hierarchical design.
 }
\label{abla:contrast}
\vspace{-10px}
\end{table}

\myheading{Results.} The results are shown in \Tref{tab:har}. MoBind achieves the best overall performance across all settings, confirming the effectiveness of its learned representations. These results demonstrate that the embedding space not only supports fine-grained cross-modal alignment but also preserves higher-level semantic structure, enabling strong class-level discriminability. The improvement further highlights the importance of the MTP auxiliary task, which regularizes the model by preventing overfitting to alignment cues and helps retain action semantics critical for recognition tasks.

\subsection{Ablation Study}

\myheading{Multi-level Contrastive Objectives.} \Tref{abla:contrast} reports an ablation study of the alignment objectives and their impact on R@1 retrieval, synchronization, and body-part localization. We propose using contrastive losses at multiple levels—token, single-sensor, and multi-sensor—and find that each contributes meaningfully to the model's overall performance. This validates our hierarchical design: fine-grained objectives capture sub-second, limb-specific cues, while the global objective aggregates full-body motion patterns, providing complementary benefits.

\myheading{Masked Token Prediction.} \Tref{abla:mtp} evaluates the impact of the Masked Token Prediction (MTP) auxiliary task on action recognition under both fine-tuning and 1-NN settings. Adding MTP helps MoBind preserve class-level semantics, yielding consistent gains across all settings and nearly 20\% improvement on TotalCapture for both classifiers. This highlights the regularizing effect of MTP in retaining semantic information beneficial for downstream action recognition, while still maintaining MoBind’s strong fine-grained temporal alignment performance.

\begin{figure}[t]
\centering
\includegraphics[width=\linewidth]{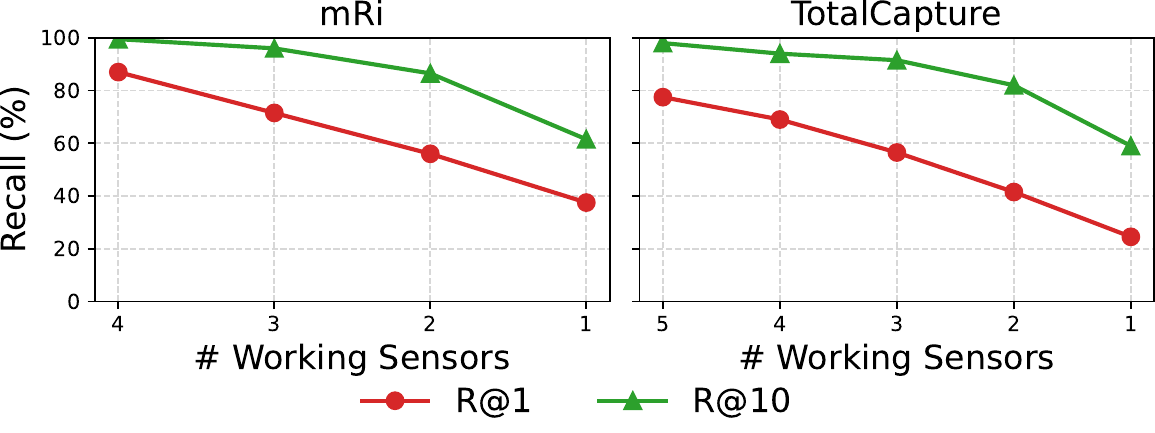}
\vskip -0.1in
\caption{
{\bf Robustness to Sensor Failure.} Retrieval performance (R@1 and R@5) under different sensor availability conditions. R@k measures the percentage of queries for which the ground-truth video appears within the top-\(k\) retrieved results, given the representation computed from a subset of IMU sensors. In general, using more IMUs provides a more complete motion representation and thus improves retrieval accuracy. MoBind remains highly effective even when some sensors are unavailable, demonstrating strong performance under partial sensor input and highlighting its robustness for real-world deployment.
}
\label{fig:drop}
\end{figure}

\setlength{\tabcolsep}{3pt}
\begin{table}[]
\centering
\begin{tabular}{lcccc}
\toprule
\multirow{2}{*}[-1ex]{} & \multicolumn{2}{c}{mRi} & \multicolumn{2}{c}{TotalCapture}\\ 
\cmidrule(lr){2-3} \cmidrule(lr){4-5}
& \multicolumn{1}{c}{Finetune} &  \multicolumn{1}{c}{1-NN} & \multicolumn{1}{c}{Finetune} & \multicolumn{1}{c}{1-NN}\\
\midrule
MoBind w/o MTP & 0.97 & 0.76 & 0.55 & 0.53\\
MoBind        & \textbf{0.98} & \textbf{0.86} & \textbf{0.72} & \textbf{0.71}\\
\bottomrule 
\end{tabular}
\vskip -0.1in
\caption{
 \textbf{Effect of Masked Token Prediction on model performance.} The MTP significantly improves action recognition on both datasets, demonstrating its importance for retaining action-level semantics.}
\label{abla:mtp}
\vspace{-10px}
\end{table}

\myheading{Robustness to Sensor Failure.}
MoBind’s modular design enables it to operate effectively with any number of IMU sensors, maintaining functionality even when some sensors become unavailable. To prepare for and evaluate such failure scenarios, we train a single model using complete sensor data augmented with random sensor dropouts. At test time, we simulate sensor failure by randomly masking out subsets of sensors and measuring IMU-to-video retrieval accuracy. As shown in~\Fref{fig:drop}, while performance naturally degrades with fewer sensors, it remains strong and continues to outperform or match baselines that rely on full sensor input. This highlights the strength of our design, which enables robust operation under partial sensor availability—an essential feature for real-world deployment.

\section{Conclusion}
MoBind is a hierarchical contrastive framework for aligning IMU signals with video-based skeletal motion. It addresses prior limitations by (i) focusing IMUs on motion-relevant pose cues instead of raw pixels, (ii) modeling multi-sensor structure by aligning each body part with its corresponding IMU, and (iii) enforcing alignment at token, local, and global levels. This representation supports several downstream tasks and achieves state-of-the-art IMU–video alignment on mRi, TotalCapture, and EgoHumans while retaining class semantics via the MTP auxiliary task.

{\small \myheading{Acknowledgment:} This work was funded by the Australian Institute for Machine Learning (Adelaide University) and the Centre for Augmented Reasoning, an initiative by the Department of Education, Australian Government.}

{
    \small
    \bibliographystyle{ieeenat_fullname}
    \bibliography{longstrings, main}

@string{aaai = "Proceedings of AAAI Conference on Artificial Intelligence"}

@string{bmvc =  "Proceedings of the British Machine Vision Conference"}

@string{cvpr   = "Proceedings of the {IEEE} Conference on Computer Vision and Pattern Recognition"}

@string{eccv  = "Proceedings of the European Conference on Computer Vision"}

@string{fg = "Proceedings of the International Conference on Automatic Face and Gesture Recognition"}

@string{icassp="Proceedings of the IEEE International Conference on Acoustics, Speech and Signal Processing"}

@string{iccv =  "Proceedings of the International Conference on Computer Vision"}

@string{icml =  "Proceedings of the International Conference on Machine Learning"}

@string{iclr ="Proceedings of International Conference on Learning and Representation"}

@string{iros =  "Proceedings of the IEEE/RSJ Conference on Intelligent Robots and Systems"}

@string{nips = "Advances in Neural Information Processing Systems"}

@string{pami =  "IEEE Transactions on Pattern Analysis and Machine Intelligence"}

@string{wacv =  "Proceedings of the IEEE Workshop on Applications of Computer Vision"}

@string{imwut = "Proceedings of the ACM on Interactive, Mobile, Wearable and Ubiquitous Technologies"}

@inproceedings{Xu2024FineParserAF,
  title={FineParser: A Fine-grained Spatio-temporal Action Parser for Human-centric Action Quality Assessment},
  author={Jinglin Xu and Sibo Yin and Guohao Zhao and Zishuo Wang and Yuxin Peng},
  year={2024},
  booktitle={IEEE/CVF Conference on Computer Vision and Pattern Recognition}  
}

@inproceedings{mtlaqa,
  title={What and How Well You Performed? A Multitask Learning Approach to Action Quality Assessment},
  author={Parmar, Paritosh and Tran Morris, Brendan},
  booktitle={IEEE/CVF Conference on Computer Vision and Pattern Recognition},
  year={2019}
}

@INPROCEEDINGS{yifeng_exrac_aaai_2024, 
    title={Count What You Want: Exemplar Identification and Few-Shot Counting of Human Actions in the Wild}, 
    author={Huang, Yifeng and Nguyen, Duc Duy and Nguyen, Lam and Pham, Cuong and Hoai, Minh},
    booktitle=aaai,
    year={2024}}

@inproceedings{m_Wang-Hoai-FG18,
 author = {Boyu Wang and Minh Hoai},
 title = {Predicting Body Movement and Recognizing Actions: an Integrated Framework for Mutual Benefits},
 year = {2018},
 booktitle = fg,
 doi = {10.1109/FG.2018.00056},
}

@inproceedings{m_Wang-etal-CVPR20,
 author = {Boyu Wang and Lihan Huang and Minh Hoai},
 title = {Active Vision for Early Recognition of Human Actions},
 year = {2020},
 booktitle = cvpr,
 doi = {10.1109/CVPR42600.2020.00116},
}

@ARTICLE{duc_caracount_tpami_2025,
  author={Nguyen, Duc Duy and Nguyen, Lam Thanh and Huang, Yifeng and Pham, Cuong and Hoai, Minh},
  journal=pami, 
  title={Class-Agnostic Repetitive Action Counting Using Wearable Devices}, 
  year={2025},
  volume={47},
  number={6}}

@INPROCEEDINGS{9207296,
  author={Li, Ying and Wang, Chenxi and Cao, Yu and Liu, Benyuan and Tan, Joanna and Luo, Yan},
  booktitle={International Joint Conference on Neural Networks}, 
  title={Human pose estimation based in-home lower body rehabilitation system}, 
  year={2020},
}

@InProceedings{Ahad_2019_CVPR_Workshops,
author = {Atiqur Rahman Ahad, Md and Das Antar, Anindya and Shahid, Omar},
title = {Vision-based Action Understanding for Assistive Healthcare: A Short Review},
booktitle = cvprw,
year = {2019}
}

@inproceedings{10.1145/2556288.2557116,
author = {Morris, Dan and Saponas, T. Scott and Guillory, Andrew and Kelner, Ilya},
title = {RecoFit: using a wearable sensor to find, recognize, and count repetitive exercises},
year = {2014},
booktitle = {Proceedings of the SIGCHI Conference on Human Factors in Computing Systems},
}

@InProceedings{Khirodkar_2023_ICCV,
    author    = {Khirodkar, Rawal and Bansal, Aayush and Ma, Lingni and Newcombe, Richard and Vo, Minh and Kitani, Kris},
    title     = {Ego-Humans: An Ego-Centric 3D Multi-Human Benchmark},
    booktitle = iccv,
    year      = {2023},
}

@inproceedings{10.5555/3327757.3327874,
author = {Korbar, Bruno and Tran, Du and Torresani, Lorenzo},
title = {Cooperative learning of audio and video models from self-supervised synchronization},
year = {2018},
booktitle = nips,
}

@InProceedings{Chung16a,
  author       = {Joon~Son Chung and Andrew Zisserman},
  title        = "Out of time: automated lip sync in the wild",
  booktitle    = "Asian Conference on Computer Vision Workshop",
  year         = {2016},
}

@InProceedings{Fernandez-Labrador_2024_CVPR,
    author    = {Fernandez-Labrador, Clara and Ak\c{c}ay, Mertcan and Abecassis, Eitan and Massich, Joan and Schroers, Christopher},
    title     = {DiVAS: Video and Audio Synchronization with Dynamic Frame Rates},
    booktitle = cvpr,
    year      = {2024},
}

@inproceedings{
zhang2024unimts,
title={Uni{MTS}: Unified Pre-training for Motion Time Series},
author={Xiyuan Zhang and Diyan Teng and Ranak Roy Chowdhury and Shuheng Li and Dezhi Hong and Rajesh K. Gupta and Jingbo Shang},
booktitle=nips,
year={2024},
}

@inproceedings{
tan2023egodistill,
title={EgoDistill: Egocentric Head Motion Distillation for Efficient Video Understanding},
author={Shuhan Tan and Tushar Nagarajan and Kristen Grauman},
booktitle=nips,
year={2023},
}

@inproceedings{das2025primus,
  title={PRIMUS: Pretraining IMU Encoders with Multimodal Self-Supervision},
  author={Das, Arnav M and Tang, Chi Ian and Kawsar, Fahim and Malekzadeh, Mohammad},
  booktitle=icassp,
  year={2025},
}

@inproceedings{zhang2025masked,
  title={Masked video and body-worn IMU autoencoder for egocentric action recognition},
  author={Zhang, Mingfang and Huang, Yifei and Liu, Ruicong and Sato, Yoichi},
  booktitle=eccv,
  year={2025},
}

@InProceedings{Afouras20b,
     author       = {Triantafyllos Afouras and Andrew Owens and Joon~Son Chung and Andrew Zisserman},
     title        = {Self-Supervised Learning of Audio-Visual Objects from Video},
     booktitle    = eccv,
     year         = {2020},
    }

@inproceedings{10.1145/3485730.3485937,
author = {Xu, Huatao and Zhou, Pengfei and Tan, Rui and Li, Mo and Shen, Guobin},
title = {LIMU-BERT: Unleashing the Potential of Unlabeled Data for IMU Sensing Applications},
year = {2021},
booktitle = {Proceedings of the 19th ACM Conference on Embedded Networked Sensor Systems},
}

@article{haresamudram2021contrastive,
  title={Contrastive predictive coding for human activity recognition},
  author={Haresamudram, Harish and Essa, Irfan and Pl{\"o}tz, Thomas},
  journal=imwut,
  volume={5},
  number={2},
  year={2021},
}

@article{10.1145/3699744,
author = {Dai, Gaole and Xu, Huatao and Yoon, Hyungjun and Li, Mo and Tan, Rui and Lee, Sung-Ju},
title = {ContrastSense: Domain-invariant Contrastive Learning for In-the-Wild Wearable Sensing},
year = {2024},
volume = {8},
number = {4},
journal = imwut,
}

@inproceedings{NIPS2000_9f699296,
 author = {Slaney, Malcolm and Covell, Michele},
 booktitle = nips,
 title = {FaceSync: A Linear Operator for Measuring Synchronization of Video Facial Images and Audio Tracks},
 year = {2000}
}

@inproceedings{marcheret2015detecting,
  title={Detecting audio-visual synchrony using deep neural networks.},
  author={Marcheret, Etienne and Potamianos, Gerasimos and Vopicka, Josef and Goel, Vaibhava},
  booktitle={Interspeech},
  year={2015}
}

@inproceedings{chung2019perfect,
  title={Perfect match: Improved cross-modal embeddings for audio-visual synchronisation},
  author={Chung, Soo-Whan and Chung, Joon Son and Kang, Hong-Goo},
  booktitle=icassp,
  year={2019},
}

@InProceedings{chen2021audio,
  title={Audio-visual synchronisation in the wild},
  author={Chen, Honglie and Xie, Weidi and Afouras, Triantafyllos and Nagrani, Arsha and Vedaldi, Andrea and Zisserman, Andrew},
  booktitle=bmvc,
  year={2021}
}

@article{10.1145/3411824,
author = {Zhang, Yun C. and Zhang, Shibo and Liu, Miao and Daly, Elyse and Battalio, Samuel and Kumar, Santosh and Spring, Bonnie and Rehg, James M. and Alshurafa, Nabil},
title = {SyncWISE: Window Induced Shift Estimation for Synchronization of Video and Accelerometry from Wearable Sensors},
year = {2020},
volume = {4},
number = {3},
journal = imwut,
}

@INPROCEEDINGS{9423336,
  author={Shahid, Muhammad and Beyan, Cigdem and Murino, Vittorio},
  booktitle=wacv, 
  title={S-VVAD: Visual Voice Activity Detection by Motion Segmentation}, 
  year={2021},
}

@inproceedings{10.1145/3474085.3475587,
author = {Tao, Ruijie and Pan, Zexu and Das, Rohan Kumar and Qian, Xinyuan and Shou, Mike Zheng and Li, Haizhou},
title = {Is Someone Speaking? Exploring Long-term Temporal Features for Audio-visual Active Speaker Detection},
year = {2021},
booktitle = {Proceedings of the ACM International Conference on Multimedia},
}

@InProceedings{Henschel_2019_CVPR_Workshops,
author = {Henschel, Roberto and von Marcard, Timo and Rosenhahn, Bodo},
title = {Simultaneous Identification and Tracking of Multiple People Using Video and IMUs},
booktitle = cvprw,
year = {2019}
}

@inproceedings{10.1109/IROS45743.2020.9341739,
author = {Sun, Xi and Weng, Xinshuo and Kitani, Kris},
title = {When We First Met: Visual-Inertial Person Localization for Co-Robot Rendezvous},
year = {2020},
booktitle = iros,
}

@INPROCEEDINGS{9826015,
  author={Liu, Hansi and Alali, Abrar and Ibrahim, Mohamed and Cao, Bryan Bo and Meegan, Nicholas and Li, Hongyu and Gruteser, Marco and Jain, Shubham and Dana, Kristin and Ashok, Ashwin and Cheng, Bin and Lu, Hongsheng},
  booktitle={ACM/IEEE International Conference on Information Processing in Sensor Networks}, 
  title={Vi-Fi: Associating Moving Subjects across Vision and Wireless Sensors}, 
  year={2022},
}

@article{Oord2018RepresentationLW,
  title={Representation Learning with Contrastive Predictive Coding},
  author={A{\"a}ron van den Oord and Yazhe Li and Oriol Vinyals},
  journal={ArXiv},
  year={2018},
}

@inproceedings{10.1145/3474085.3475307,
author = {Thoker, Fida Mohammad and Doughty, Hazel and Snoek, Cees G. M.},
title = {Skeleton-Contrastive 3D Action Representation Learning},
year = {2021},
booktitle = {Proceedings of the 29th ACM International Conference on Multimedia},
}

@inproceedings{Guo_Liu_Chen_Liu_Wang_Ding_2022, 
title={Contrastive Learning from Extremely Augmented Skeleton Sequences for Self-Supervised Action Recognition},  
booktitle=aaai, 
author={Guo, Tianyu and Liu, Hong and Chen, Zhan and Liu, Mengyuan and Wang, Tao and Ding, Runwei}, 
year={2022}}

@inproceedings{10.1609/aaai.v37i1.25127,
author = {Dong, Jianfeng and Sun, Shengkai and Liu, Zhonglin and Chen, Shujie and Liu, Baolong and Wang, Xun},
title = {Hierarchical contrast for unsupervised skeleton-based action representation learning},
year = {2023},
booktitle = aaai,
}

@inproceedings{
    an2022mri,
    title={m{RI}: Multi-modal 3D Human Pose Estimation Dataset using mmWave, {RGB}-D, and Inertial Sensors},
    author={Sizhe An and Yin Li and Umit Ogras},
    booktitle={Conference on Neural Information Processing Systems Datasets and Benchmarks Track},
    year={2022},
    }

@inproceedings{Trumble:BMVC:2017,
	AUTHOR = "Trumble, Matt and Gilbert, Andrew and Malleson, Charles and  Hilton, Adrian and Collomosse, John",
	TITLE = "Total Capture: 3D Human Pose Estimation Fusing Video and Inertial Sensors",
	BOOKTITLE = bmvc,
	YEAR = "2017",}

@inproceedings{girdhar2023imagebind,
  title={ImageBind: One Embedding Space To Bind Them All},
  author={Girdhar, Rohit and El-Nouby, Alaaeldin and Liu, Zhuang
and Singh, Mannat and Alwala, Kalyan Vasudev and Joulin, Armand and Misra, Ishan},
  booktitle= cvpr,
  year={2023}
}

@InProceedings{pmlr-v139-radford21a,
  title = 	 {Learning Transferable Visual Models From Natural Language Supervision},
  author =       {Radford, Alec and Kim, Jong Wook and Hallacy, Chris and Ramesh, Aditya and Goh, Gabriel and Agarwal, Sandhini and Sastry, Girish and Askell, Amanda and Mishkin, Pamela and Clark, Jack and Krueger, Gretchen and Sutskever, Ilya},
  booktitle = 	 icml,
  year = 	 {2021},
}

@article{moon2022imu2clip,
  title={IMU2CLIP: Multimodal Contrastive Learning for IMU Motion Sensors from Egocentric Videos and Text},
  author={Moon, Seungwhan and Madotto, Andrea and Lin, Zhaojiang and Dirafzoon, Alireza and Saraf, Aparajita and Bearman, Amy and Damavandi, Babak},
  journal={arXiv preprint arXiv:2210.14395},
  year={2022}
}

@inproceedings{
    despite2025,
    title={DeSPITE: Exploring Contrastive Deep Skeleton-Pointcloud-IMU-Text Embeddings for Advanced Point Cloud Human Activity Understanding},
    author={Thomas Kreutz and Max Mühlhäuser and Alejandro Sanchez Guinea},
    booktitle=iccv,
    year={2025},
    }

@InProceedings{sparse2022iashin,
  title={Sparse in Space and Time: Audio-visual Synchronisation with Trainable Selectors},
  author={Iashin, V. and Xie, W. and Rahtu, E. and Zisserman, A.},
  booktitle=bmvc,
  year={2022}
}

@InProceedings{synchformer2024iashin,
  title={Synchformer: Efficient Synchronization from Sparse Cues},
  author={Iashin, V. and Xie, W. and Rahtu, E. and Zisserman, A.},
  booktitle=icassp,
  year={2024},
}

@misc{https://doi.org/10.48550/arxiv.2303.07399,
  author = {Jiang, Tao and Lu, Peng and Zhang, Li and Ma, Ningsheng and Han, Rui and Lyu, Chengqi and Li, Yining and Chen, Kai},
  title = {RTMPose: Real-Time Multi-Person Pose Estimation based on MMPose},
  publisher = {arXiv},
  year = {2023},
}

@misc{mmpose2020,
    title={OpenMMLab Pose Estimation Toolbox and Benchmark},
    author={MMPose Contributors},
    howpublished = {\url{https://github.com/open-mmlab/mmpose}},
    year={2020}
}

@inproceedings{Kingma2015AdamAM,
  title={Adam: A Method for Stochastic Optimization},
  author={Diederik P. Kingma and Jimmy Ba},
  booktitle=iclr,
  year={2015}
}
}


\end{document}